%% file: main.tex
\documentclass{IEEEtaes}

\usepackage{color,array,amsthm}
\usepackage{graphicx}

\jvol{XX}
\jnum{XX}
\jmonth{XXXXX}
\paper{1234567}
\pubyear{2020}
\doiinfo{TAES.2020.Doi Number}

\usepackage{amsmath}
\usepackage{amsfonts}
\usepackage{multirow}
\usepackage{listings}
\usepackage[most]{tcolorbox}
\usepackage[T1]{fontenc}
\usepackage{textcomp}
\usepackage{arydshln}
\usepackage{makecell}
\usepackage{subcaption}

\lstdefinestyle{ieeeprompt}{
  basicstyle=\ttfamily\scriptsize,
  columns=fullflexible,
  breaklines=true,
  breakatwhitespace=true,
  keepspaces=true,
  showstringspaces=false,
  upquote=true,
  aboveskip=0.5\baselineskip,
  belowskip=0.5\baselineskip
}

\newtcblisting{promptbox}[2][]{%
  enhanced,
  arc=1mm,
  breakable,
  boxrule=0.4pt,
  colback=white,
  colframe=black!95,
  titlerule=0pt,
  toptitle=0mm,
  bottomtitle=0mm,
  top=0.6mm,
  left=1.2mm,right=1.2mm,top=0.8mm,bottom=0.8mm,
  listing only,
  listing options={style=ieeeprompt},
  title=\textbf{#2},
  fonttitle=\footnotesize,
  #1
}

\begin{document}

\title{OrbitTAMP: Grounding Language Models for Task and Motion Planning in Spacecraft Rendezvous\vspace{-1.45\baselineskip}}

\author{Yuji Takubo}
\member{Student Member, IEEE}
\affil{Stanford University, Stanford, CA 94305, USA}

\author{Daniele Gammelli}
\member{Member, IEEE}
\affil{Stanford University, Stanford, CA 94305, USA \\ Italian Institute of Artificial Intelligence (AI4I), Turin, Italy}

\author{Marco Pavone}
\member{Senior Member, IEEE}
\affil{Stanford University, Stanford, CA 94305, USA \\ NVIDIA Research, Santa Clara, CA 95051, USA}

\author{Simone D'Amico}
\member{Member, IEEE}
\affil{Stanford University, Stanford, CA 94305, USA}

\receiveddate{Manuscript received XXXXX 00, 0000; revised XXXXX 00, 0000; accepted XXXXX 00, 0000.\\
Y. Takubo acknowledges the Ezoe Memorial Recruit Foundation for the financial support.  }

\corresp{{\itshape (Corresponding author: Y. Takubo)}.}

\authoraddress{
Y. Takubo, D. Gammelli, M. Pavone, and S. D'Amico are with the Department of Aeronautics and Astronautics.
(e-mail: {\{ytakubo, gammelli, pavone, damicos\}@stanford.edu}).
}

\supplementary{Color versions of one or more of the figures in this article are available online at \href{http://ieeexplore.ieee.org}{http://ieeexplore.ieee.org}.}

\markboth{TAKUBO ET AL.}{OrbitTAMP: Grounding Language Models for Task and Motion Planning in Spacecraft Rendezvous }
\maketitle

\begin{abstract}
Spacecraft rendezvous and proximity operations (RPO) are currently planned through an expertise-intensive process in which engineers translate high-level operational intent into safe, dynamically feasible trajectories, creating a bottleneck to scalable operations. 
Large language model (LLM)-based agents could offer an intuitive interface for this process, although their outputs are not inherently grounded in orbital dynamics, operational constraints, or the structure of admissible spacecraft maneuvers. 
To exploit their semantic reasoning while ensuring the generated plan's physical validity, this paper presents a hierarchical framework for spacecraft task-and-motion planning (TAMP) that grounds LLM reasoning in a graph of reusable behaviors and domain-specific planning modules. 
Within this framework, a pretrained LLM maps a natural-language command to a partial mission specification. The associated planners then resolve unspecified decisions within the admissible operational space. 
Finally, trajectory optimization converts the completed mission specification into a dynamically feasible trajectory. 
Numerical experiments demonstrate that this architecture substantially improves intent recovery over direct LLM generation, achieving 98\% exact recovery of partial mission specifications across all evaluated splits when backed by frontier LLMs. 
Additional test-time-compute experiments show that, for a compact 9B model, verifier-guided revision increases exact recovery from 75\% to 88\%, while broader behavior-plan search independently improves selection among admissible trajectory realizations.
Overall, these results establish a scalable and auditable foundation for language-driven agentic planning of spacecraft RPO.
\end{abstract}


\input{introduction}

\input{related_work}

\input{method}
\input{case_study}
\input{results}

\section{Conclusion} \label{sec:conclusion}

This paper introduced a hierarchical TAMP architecture that grounds LLM reasoning in spacecraft-specific behavior abstractions, operational constraints, and orbital dynamics to enable language-commanded trajectory generation for rendezvous and proximity operations.
Through a structured mission representation, a graph of reusable spacecraft behaviors, and domain-specific planning modules, the architecture progressively translates high-level operator intent into a complete, verifiable mission specification and, ultimately, a trajectory satisfying the adopted dynamics and safety constraints.
This hierarchy separates semantic grounding, completion of unspecified mission decisions, and physical trajectory verification while preserving operator-provided information throughout the planning process.
Numerical experiments demonstrated the advantages of the hierarchical architecture over one-shot LLM-based planning, characterized how structured intent recognition and end-to-end mission satisfaction vary with model capability, and showed how test-time computation at different layers can improve semantic reliability and realized trajectory quality.
Together, these results establish the proposed framework as a step toward scalable, auditable human--spacecraft interfaces for future agentic spacecraft operations.
Future work will extend the framework to multi-spacecraft coordination and closed-loop execution with online replanning.

\section*{Acknowledgment}
AI tools were used for brainstorming, coding assistance, and language polishing.

\bibliographystyle{ieeetr}
\bibliography{main}

\input{bio}

\end{document}


\maketitle
\thispagestyle{empty}
\pagestyle{empty}

\renewcommand{\theequation}{A\arabic{equation}}
\setcounter{equation}{0}

\subsection{Learning-based Waypoint Generation} \label{app:learned_wyp_model}

This subsection discusses the detailed implementation of the learning-based waypoint generation module. 

First, training data are generated by solving the trajectory optimization problem (cf. Eq.~11) for randomized graph-valid behavior-duration sequences, scenarios, and waypoint sampling within each waypoint domain. 
In particular, the waypoint for these rows is drawn from an 11-point grid spanning each domain; the initial waypoint $\boldsymbol{w}_0$ for these rows is drawn from the same on-grid domain sampler.
A SCP solve optimizes $(\delta\lambda,\delta e_y,\delta i_y)$ jointly within the selected waypoint domains.
In this case study, 20,000 data rows are generated, and the optimized waypoint sequences from converged rows form the imitation targets.
The rows are split randomly, with 80\%/10\%/10\% used for training, validation, and testing.

The model architecture adopts a multilayer perceptron (MLP) with three hidden layers of 256 units each.
In order to account for the disconnected waypoint domains, each disconnected coordinate is represented by a magnitude logit and a binary sign. 
Namely, at each phase $m = 1, ..., M$, it outputs Gaussian means and standard deviations for the three continuous variables and logits for the two signs:
\begin{subequations}\label{eq:wyp_reparameterization}
\small
\begin{align}
z_{\lambda,k}
&=\operatorname{logit}\!\left(
\frac{a\delta\lambda_k-\ell_{\lambda,k}}
     {u_{\lambda,k}-\ell_{\lambda,k}}\right), \ z_{v,k}
=\operatorname{logit}\!\left(
\frac{|a\delta v_k|-\ell_{v,k}}
     {u_{v,k}-\ell_{v,k}}\right),\\
s_{v,k}
&=\mathbb{I}[a\delta v_k\geq0],
\qquad v\in\{e_y,i_y\}.
\end{align}
\end{subequations}
Here, $[\ell_{\lambda,k},u_{\lambda,k}]$ is the connected along-track interval, and $[\ell_{v,k},u_{v,k}]$ is the admissible magnitude interval.
The inverse map uses an affine sigmoid, which guarantees that the prediction stays inside the graph-prescribed domain.

The model is trained using unweighted imitation learning, supervised on the converged rows.
Each training example already contains one sampled graph-valid behavior-duration sequence $(\boldsymbol{b}_{1:M},\boldsymbol{d}_{1:M})$ and its optimized waypoint targets $\boldsymbol{w}_{1:M}^{\star}$.
The loss uses Gaussian negative log-likelihood (NLL) for the continuous targets and binary cross-entropy (BCE) for the sign targets.
The per-example loss and unweighted minibatch objective are
\begin{equation}\label{eq:wyp_loss}
\small
\begin{aligned}
\mathcal{L}_{\phi}
& =
\frac{1}{|\mathcal{I}|} \sum_{i\in\mathcal{I}}\ell_{\phi}^{(i)}. \\
\ell_{\phi}
&=
\frac{1}{3M}
\sum_{k=1}^{M}\sum_{c=1}^{3}
\operatorname{NLL}_{\mathcal{N}}\!\left(
\widetilde{z}_{kc};
\mu_{\phi,kc},\sigma_{\phi,kc}
\right)
\\
& +
\frac{1}{2M}
\sum_{k=1}^{M}\sum_{c=1}^{2}
\operatorname{BCE}\!\left(
s_{kc};\pi_{\phi,kc}
\right),
\end{aligned}
\end{equation}
where $\mathcal{I}$ is the minibatch index set, and each component $\widetilde{z}_{kc}=(z_{kc}-\bar z_{kc})/\tau_{kc}$ is normalized using the training-set mean $\bar z_{kc}$ and standard deviation $\tau_{kc}$.
The quantities $\mu_{\phi,kc}$ and $\sigma_{\phi,kc}$ are the corresponding predicted Gaussian parameters, and $\pi_{\phi,kc}$ is the softmax probability of the positive-sign class.
The factors $3M$ and $2M$ average the continuous and sign losses over the $M$ selected phases, so two- and three-phase data receive equal weight.

\subsection{Learning-based Behavior-Duration Planning} \label{app:learned_behav_model}

This subsection discusses the detailed implementation of the learning-based behavior-duration planning module. 

First, the dataset generation proceeds similarly to learning-based waypoint generation, where data are generated by solving Eq.~11 for randomized graph-valid behavior-duration sequences and scenarios.
However, the waypoint corresponding to the sampled behavior-duration pair is selected using the corresponding waypoint policy $\pi_w$. 
Namely, the trained model $Q_\psi$ is strictly dependent on how the waypoint is placed. 

The metric surrogate model is formulated as an MLP and a gated recurrent unit (GRU) with a 128-dimensional hidden state.
The final output comprises a scalar convergence prediction $\hat{p}_{\mathrm{conv}} \in [0,1]$ and a metric prediction $\hat{\tilde{\boldsymbol{q}}} \in \mathbb{R}^{N_\mathcal{P}}$.
To predict the convergence of the subsequent optimization, both converged and nonconverged rows are used for training.

The loss combines a BCE term on predicted convergence with a Huber loss over the realized metrics; the latter is evaluated only on converged rows:
\begin{equation} \label{eq:loss_behavior}
\small
\begin{aligned}
    \mathcal{L}_{\psi}
    =\ &
    \frac{1}{|\mathcal{I}|}\sum_{i\in\mathcal{I}}
    \operatorname{BCE}\!\left(\hat p^{(i)}_{\mathrm{conv}}, y^{(i)}_{\mathrm{conv}}\right)
    \\
    +\ &
    \frac{\lambda_q}{N_{\mathcal{P}}|\mathcal{I}_{conv}|}
    \displaystyle\sum_{i\in\mathcal{I}_{conv}}\
    \displaystyle\sum_{j=1}^{N_{\mathcal{P}}} \ \!\!\operatorname{Huber}_1\!\left(\hat{\tilde q}^{(i)}_{j}-\tilde q^{(i)}_{j}\right),
\end{aligned}
\end{equation}
where $\mathcal{I}_{conv}=\{i\in\mathcal{I}:y^{(i)}_{\mathrm{conv}}=1\}$ denotes the converged rows, where $y_{\mathrm{conv}} \in \{0,1\}$ is the binary indicator of numerical convergence in the optimization, and $\lambda_q$ is a weighting parameter.
The value $\tilde q_{j}$ is normalized using the median and interquartile range (IQR) of metric $q_{j}$ computed from the converged training rows, and $\hat{\tilde q}_{j}$ is its prediction.
For the safety margin, the raw value is clipped at $25\,\mathrm m$ before scaling to improve prediction near the boundary.
$\operatorname{Huber}_1$ is the Huber loss with unit transition width, which prevents a few large residuals from dominating the gradient.
Thus, the convergence head uses every row, whereas the metric loss uses only converged rows.
Optimization convergence probabilities are Platt-calibrated on the validation set.
In this case study, 20,000 data rows are generated, and the dataset is partitioned randomly into an 80\%/10\%/10\% split.

At deployment, all graph-valid behavior-duration candidates consistent with $\mathcal{M}_0$ are enumerated (cf. Table~II) and evaluated by $Q_\psi$ in one batched forward pass.
Candidates with predicted convergence probability below $p_{\min} =0.5$ are removed, and the remainder are ranked by $\rho_{\mathcal{P}}$ to determine the top-ranked behavior-duration plan.

\subsection{Prompt Structure} \label{app:parser_prompt}
The following box summarizes the prompt supplied to the LLM rather than reproducing it verbatim.
The behavior and waypoint ontology included in the prompt is the same as that defined in Tables~I and~II.
In the actual implementation, this template is augmented with domain-specific interpretation rules, summarized by the parsing principles below, and scenario-specific grounding based on the initial waypoint domain.

\begin{promptbox}{Condensed initial parsing prompt}
System:
Parse the command into a partial MissionConfig. Preserve explicit operator intent, leave unsupported fields unspecified, and return one JSON object without prose.

User:
Output schema:
{
  "preferences": {"priority": [...]},
  "hard_constraints": {...},
  "b_seq": [...], "d_seq": [...],
  "tof_orbit": null,
  "tof_bound": {"lower_orbit": null,
                "upper_orbit": null},
  "x_seq": [...]
}

Parsing principles:
1. Extract only stated intent; do not complete the mission.
2. Preserve the number and order of stated phases.
3. For a multi-phase command, represent each stated phase with corresponding behavior, duration, and waypoint entries, using unspecified values when information is omitted.
4. Distinguish where a maneuver starts from where the operator requests it to end.
5. Do not add intermediate phases or behaviors that the operator did not state.
6. Use only the supplied ontology, metrics, and constraints.

<behavior and waypoint ontology>
start_domain = <domain>
operator_command = <command>
\end{promptbox}

If verification fails, the following message is appended to the original conversation, so the model retains the schema, ontology, command, and previous response.

\begin{promptbox}{Verifier-guided revision message}
Previous response:
<candidate MissionConfig>

Verifier feedback:
<list of deterministic errors>

Return a complete corrected MissionConfig as JSON. Preserve fields not implicated by the errors and do not add unstated mission intent.
\end{promptbox}

\subsection{Command-Benchmark Construction}
\label{app:command_benchmark}

Four linguistic styles are used to render the operator commands: formal, natural, concise, and radio.
Table~\ref{tab:language_style_examples} illustrates these styles using the same underlying reference mission intent $\mathcal{M}_0^\star$.
The linguistic realization changes across styles while preserving the supported mission semantics and behavior--waypoint ontology.

Table~\ref{tab:command_splits} summarizes their use across the three benchmark splits.
The development and held-out-instance splits use the formal, natural, and concise styles, with the latter containing independently-sampled mission specifications.
The held-out-structural split also uses independently sampled mission specifications, but renders them exclusively in the unseen radio register to introduce a linguistic-structure shift.

\begin{table}[t]
    \centering
    \caption{Illustrative renderings of the same reference mission intent
    across the supported language styles. }
    \label{tab:language_style_examples}
    \small
    \setlength{\tabcolsep}{4pt}
    \renewcommand{\arraystretch}{1.03}
    \begin{tabular}{l p{0.8\columnwidth}}
        \hline
        {Style} & {Command realization} \\
        \hline
        Formal &
        Circumnavigate the target through the near-target operating region.
        Prioritize maximizing observation quality over minimizing completion time. \\

        Natural &
        Move the chaser around the target along a circumnavigation path.
        Give observation quality higher priority than completion time. \\

        Concise &
        Chaser task: target circumnavigation.
        Priority order: observation quality, then completion time. \\

        Radio &
        Circumnavigate target, near-field.
        Prioritize observation quality over completion time, all phases. \\
        \hline
    \end{tabular}
\end{table}

\begin{table}[t]
    \centering
    \caption{Command-benchmark splits used for language-conditioned evaluation.}
    \label{tab:command_splits}
    \footnotesize
    \setlength{\tabcolsep}{4pt}
    \renewcommand{\arraystretch}{1.08}
    \begin{tabular}{
        >{\centering\arraybackslash}m{0.21\columnwidth}
        p{0.25\columnwidth}
        p{0.40\columnwidth}
    }
        \hline
        {Split}
        & {Mission instances}
        & {Language styles} \\
        \hline

        \makecell[c]{Development}
        & Development sample
        & Formal, natural, concise \\

        \makecell[c]{Held-out \\ instance}
        & Independently sampled
        & Formal, natural, concise (seen) \\

        \makecell[c]{Held-out \\ structural}
        & Independently sampled
        & Radio register (unseen) \\

        \hline
    \end{tabular}
\end{table}

\subsection{Verification of $\mathcal{M}_0$}
\label{app:m0_verification}
Each $\mathcal{M}_0$ proposed during intent parsing is checked before acceptance. It must contain the required \texttt{MissionConfig} fields with the correct data types, use only the behaviors, waypoint regions, metrics, and constraints defined in this paper, and contain valid and consistent time bounds. Relative-state waypoints must contain six coordinates, and the behavior, duration, and waypoint sequences must contain the same number of phase entries.

At every revision budget, a small deterministic check also compares the proposed $\mathcal{M}_0$ with certain movement sequences explicitly stated in the command. It applies only to (i) an ordered list of waypoint regions, (ii) an approach followed by departure, or (iii) a direct cross-side flyby or fast passage.
Expressions such as $F_+$, $E_0$, $F_-$, and their $V$-bar equivalents are matched to the waypoint regions in Table~I, and the behavior--waypoint graph identifies the behavior associated with each stated movement. For example, $F_+ \rightarrow E_0 \rightarrow F_-$ means that the first waypoint is in $E_0$ and the second is in $F_-$; $F_+$ is the initial region and is therefore not repeated as a waypoint. At $N_{\mathrm{rev}}=0$, any disagreement causes rejection; at $N_{\mathrm{rev}}>0$, it is returned to the LLM for revision.
Commands outside these three cases receive no movement-specific feedback, and unrelated fields may remain at $\bot$.

After intent parsing, a final check determines whether every populated behavior can connect the preceding waypoint region to the requested next region in the behavior--waypoint graph, beginning from the known initial region. If a behavior is inconsistent but the requested waypoint remains reachable through another allowed behavior, that behavior entry is reset to $\bot$, and the waypoint is preserved. 
This check does not fill unspecified fields or complete the behavior sequence.

\subsection{SCP Parameters} \label{sec:appendix_scp}

Table \ref{tab_appendix:scp_param} presents the hyperparameters of $\texttt{SCvx}\star$ based on the notation in \cite{oguri2023successive}.
All optimization variables are normalized based on the warm-start, so that all variables are scaled to lie within $[-1,1]$.

\begin{table}[ht]
\centering
\caption{Hyperparameters of $\texttt{SCvx}\star$.}
\renewcommand*{\arraystretch}{1.05}
\begin{tabular}{cc}
    \hline
    Hyperparameter & Value \\
    \hline
     $\epsilon$ & $10^{-3}$    \\
     $\{\rho_0, \rho_1, \rho_2\}$ & $\{ 0.0, 0.25, 0.7 \}$ \\
     $\{\alpha_1, \alpha_2, \beta, \gamma \}$ & $\{ 2,2,1.5,0.9\}$ \\
     $\{ r^{(1)}, r_{\min}, r_{\max} \}$ & $\{ 0.2, 10^{-9}, 10 \}$ \\
     $\{w^{(1)}, w_{\max}\}$ & $\{ 100, 10^{20} \}$ \\
     $\#$ max. iter & 100  \\
    \hline
    \end{tabular}
    \label{tab_appendix:scp_param}
\end{table}

\bibliographystyle{ieeetr}
\bibliography{main}

%% file: introduction.tex
\section{Introduction}

\begin{figure*}[t]
    \centering
    \includegraphics[width=1\textwidth]{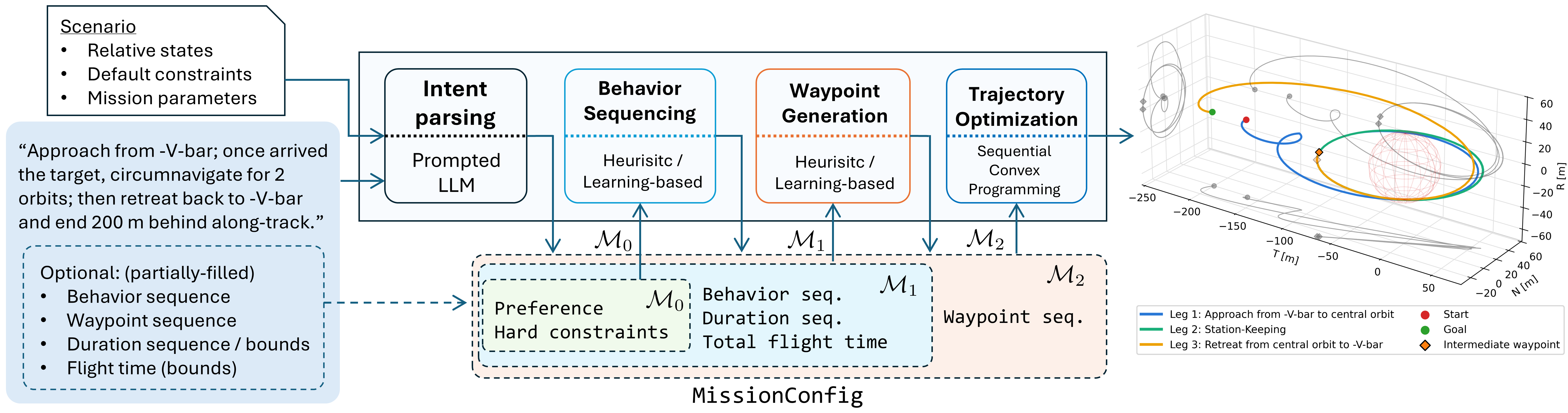}
    \caption{Overview of the proposed framework. The intent parsing initializes the partial mission specification $\mathcal{M}_0$; behavior sequencing produces $\mathcal{M}_1$; waypoint generation produces $\mathcal{M}_2$; and trajectory optimization generates the final state and control trajectories. All three specifications are successive states of the same \texttt{MissionConfig}.}
    \label{fig:rages}
\end{figure*}

The proliferation of distributed space systems, such as swarms and constellations, is increasing the need for scalable spacecraft operations. 
Rendezvous and proximity operations (RPO), in particular, require complex maneuver sequences for inspection, repair, refueling, and other servicing tasks. 
Traditionally, teams of operators and mission designers translate high-level mission intent or concepts of operations (CONOPS) into explicit trajectory specifications using their knowledge of astrodynamics, mission constraints, and available guidance modes. 
This dependence on specialized expertise limits the scalability of future spacecraft operations.

Large language models (LLMs) have the potential to establish natural language as an intuitive and broadly accessible interface between high-level mission intent and executable spacecraft trajectories.
Specifically, recent advances in agentic LLM frameworks have demonstrated that foundation models (FMs) can interpret high-level instructions, decompose objectives into multistep plans, invoke external tools, and coordinate reasoning with action across interactive decision-making, autonomous scientific experimentation, and embodied operations \cite{yao2022react, boiko2023autonomous, brohan2023saycan}.
For spacecraft RPO, however, semantic interpretation alone is insufficient. Language-model outputs must be grounded in spacecraft-specific behavior structures, operational constraints, and orbital dynamics before they can support safe execution. The central challenge is therefore not merely to translate language into a quantitative configuration, but to embed LLM reasoning within a planning architecture that converts incomplete intent into admissible and physically realizable spacecraft motion.

Consider the command:
``\textit{approach the target slowly from the $-V$-bar, circumnavigate it for two orbits, and then depart to a $50 \, \mathrm{m}$ along-track separation.}''
To a human operator, this command conveys a clear maneuver intent through an ordered composition of familiar spacecraft behaviors: approach, circumnavigate, and depart.
Such reusable behaviors provide abstractions for organizing complex motion, similar in function to \textit{skills} \cite{brohan2023saycan} or \textit{meta-actions} \cite{wang2025alpamayo} in broader robotic systems.
As a trajectory-planning specification, however, the command remains incomplete.
It dictates several explicit requirements, including the two-orbit circumnavigation and final separation, while leaving other decisions such as phase durations, intermediate waypoints, and trajectory constraints unspecified. 
This distinction is central to spacecraft trajectory planning: conventional task and trajectory planners typically operate on formally defined actions, objectives, and constraints, whereas natural-language mission commands generally specify only a subset of this information, and the unstated remainder is where mission-design expertise must be applied.
Grounding such a command therefore requires identifying the stated intent, resolving unspecified decisions within the admissible spacecraft operational space, and constructing a dynamically feasible and safe trajectory.

To address this problem, this paper introduces a hierarchical framework for grounding LLM reasoning for spacecraft task-and-motion planning (TAMP), illustrated in Fig.~\ref{fig:rages}. 
Extending the preliminary concepts of structural decomposition for spacecraft behavior planning \cite{telaar2017edeorbit, takubo2026intent, zhang2025llm}, the framework organizes the planning process into three stages: semantic grounding, completion of unspecified mission decisions, and physical trajectory verification. 
A pretrained LLM first maps the natural-language command to a partial mission specification that preserves the operator’s stated intent while explicitly representing unspecified information.
A graph of reusable spacecraft behaviors, together with domain-specific planning modules, then completes the behavior sequence, timing, and waypoint decisions within the admissible operational space.
Finally, trajectory optimization instantiates the resulting candidates as state-control trajectories and ensures their dynamical feasibility and safety. 
This hierarchical architecture grounds LLM reasoning at each stage while enabling multistage planning, computational tool use, and intermediate verification.

The central contribution of this paper is a hierarchical planning formulation in which operator underspecification is preserved and systematically completed, realized as an agentic architecture that scaffolds pretrained, off-the-shelf LLMs with structured, spacecraft-specific planning and verification modules.
Because every stage completes the same partial specification, stated intent is preserved by construction rather than by prompt compliance and remains checkable throughout, while unstated decisions are resolved by the behavior graph and planning modules under orbital dynamics and safety constraints rather than left to the LLM.
This design is motivated by both the scarcity of domain-specific training data for space applications \cite{foutter2025space-llava} and the need for transparent, verifiable, and explicitly constrained decision-making in safety-critical spacecraft operations.
The resulting architecture provides an auditable human--spacecraft interface and supports scalable language-driven planning for future distributed space systems.
Specifically, the contributions of this work are as follows:
\begin{itemize}
    \item A spacecraft RPO planning representation comprising a graph of reusable behaviors and a structured mission configuration that preserves partial operator specifications and supports hierarchical completion of behavior, timing, and waypoint decisions.
    \item An LLM-backed TAMP formulation that uses these representations to decouple intent parsing from trajectory realization, grounding semantic decisions in the admissible behavior structure and resolving unstated decisions within operational constraints and spacecraft dynamics.
    \item A stage-wise and end-to-end evaluation over 300 procedurally generated commands and multiple LLM backends and a deterministic parsing baseline that quantifies the effects of the grounding scaffold, model capability, and targeted test-time computation through verifier-guided revision and behavior-plan search.
\end{itemize}

The remainder of the paper is organized as follows.
Sec.~\ref{sec:related_work} compares the proposed framework with spacecraft planning, language grounding, and robotic TAMP.
Sec.~\ref{sec:method} defines the common mission representation, behavior graph, and systematic completion process.
Sec.~\ref{sec:case_study} describes the on-orbit inspection case study and its specific implementations.
Sec.~\ref{sec:results} reports the module-level and end-to-end evaluation results.
Finally, Sec.~\ref{sec:conclusion} concludes the paper and discusses future work.

%% file: related_work.tex
\section{Related Work}
\label{sec:related_work}

This work builds on two closely related lines of research: spacecraft planning and trajectory generation, and language-grounded TAMP for robotic systems. 
The following review for each line of research is provided below.

\subsection{Spacecraft Planning, Language Grounding, and Trajectory Optimization}

Classical spacecraft autonomy separates high-level mission planning from guidance, navigation, and control.
Onboard systems have demonstrated activity planning, scheduling, execution monitoring, and replanning from predefined goals \cite{muscettola1998remote,chien2005ase}.
Autonomous decision-making based on discrete logic has also been developed for formation flying and rendezvous and proximity operations \cite{telaar2017edeorbit,Guffanti2023Visors,kruger2024adaptive}.
In these systems, expert-defined activity models and transition logic specify high-level tasking and behavior sequencing, while waypoint and maneuver generation is generally handled by separate modules.

Trajectory optimization provides a principled mechanism for generating a state-control trajectory that enforces the underlying dynamics, waypoint conditions, and operational constraints \cite{lawden1963optimal,malyuta_scp_2022}.
Learning-based methods can improve its numerical performance through warm-starting or approximation \cite{Banerjee_2020,kim2022guided,art_ieeeaero24,takubo2024towards}.
These methods generally assume that the maneuver structure, waypoints, objectives, and constraints have already been specified based on the desired high-level spacecraft behaviors.

Recent work has explored the use of semantic reasoning for spacecraft operations \cite{carrasco2025large,foutter2025space-llava,jain2026autonomous}.
Parallel research translates natural-language safety requirements into mathematical representations, including explicit constraints and signal temporal logic formulas \cite{chen2023nl2tl,brosius2026semantic}.
Prior work by the authors learned language-conditioned trajectory warm starts from paired command--trajectory data and refined them through sequential convex programming (SCP) \cite{takubo2026language}.

Collectively, these approaches provide complementary capabilities for high-level spacecraft autonomy and continuous trajectory generation with potential for connection with semantics. 
Their interface nevertheless remains dependent on expert-specified maneuver definitions or complete trajectory-level representations, leaving no structured mechanism for preserving partial operator intent, resolving unspecified mission decisions, and verifying the resulting behavior sequence under orbital dynamics and safety constraints, in a scalable manner. 

\subsection{Learning-Guided Task and Motion Planning}

TAMP integrates discrete task-level decision-making with continuous motion generation \cite{garrett2021integrated}.
Classical TAMP evaluates symbolic choices through motion-level sampling or optimization \cite{garrett2020pddlstream,toussaint2015logic}, whereas learning-guided approaches leverage data to prioritize task decisions, propose continuous parameters, or filter infeasible candidates \cite{kim2022gtamp,xu2022neuralfeasibility,silver2021symbolic}.
These methods provide a general framework for coupling behavior selection with trajectory generation. 

Language-conditioned planning in robotics provides a complementary perspective by enabling high-level instructions to guide planning and control.
Existing approaches combine the reasoning capabilities of language models with learned affordances, executable programs, or external planners \cite{brohan2023saycan,huang2022language,liang2023code}. 
Some use LLM to propose or prioritize skill-level plans under feasibility models \cite{lin2023text2motion}, while others translate instructions into a structured formal task specifications used by downstream planners \cite{chen2024autotamp,kambhampati2024llmmodulo}.

Collectively, these methods demonstrate how language-level reasoning can be connected to executable motion through structured planning abstractions. 
However, their formulations generally seek a sufficiently specified symbolic problem or skill sequence and do not address the spacecraft-specific need to preserve operator underspecification while completing behavior, timing, and waypoint decisions within an admissible decision space for RPO missions.

Taken together, prior work in spacecraft mission planning, trajectory optimization, language grounding, and robotic TAMP provides complementary pieces towards language-driven spacecraft operations. 
The proposed framework unifies these directions by representing language as a partial mission specification and systematically completing its unspecified decisions through a graph of admissible spacecraft behaviors, trajectory-aware evaluation, and final trajectory optimization.

%% file: method.tex
\section{Methodology}
\label{sec:method}

The objective of this work is to generate an executable spacecraft trajectory from an operator’s high-level intent, expressed as a natural-language command.
The planner must first infer an admissible sequence of spacecraft behaviors and then solve for the corresponding continuous trajectory.
Because natural-language commands may omit relevant details or leave them underspecified, the framework must also identify and resolve missing information while preserving the operator’s stated intent.

The overall architecture is summarized in Fig.~\ref{fig:rages}.
The framework takes as input a natural-language command and a mission scenario specifying the initial relative state, default constraints, and other mission parameters.
A shared representation, \texttt{MissionConfig}, carries the evolving mission specification throughout the pipeline.
First, the intent parser maps the command to a partial specification, $\mathcal{M}_0$.
The behavior-duration planning stage then uses the behavior graph to complete the behavior sequence and phase durations, yielding $\mathcal{M}_1$.
Next, the waypoint planning stage completes the corresponding waypoint specifications to produce $\mathcal{M}_2$.
Finally, trajectory optimization instantiates $\mathcal{M}_2$ as state–control trajectories subject to the adopted dynamics and safety constraints.
Thus, $\mathcal{M}_0$, $\mathcal{M}_1$, and $\mathcal{M}_2$ represent successive stages in the completion of a single \texttt{MissionConfig}, with operator-specified information preserved throughout the pipeline.

The following subsection introduces the shared planning representations that support the proposed hierarchy. 
The planning modules are then described in sequence to show how these representations structure the admissible search space while preserving operator intent and ensuring that the resulting trajectory satisfies the adopted safety and dynamical feasibility requirements.

\subsection{Common Planning Representation}

To make this planning problem tractable, this paper introduces two complementary abstractions: a graph-based representation of reusable spacecraft behaviors and a structured representation of the mission specification.

\subsubsection{Behavior--Waypoint Graph}

\begin{figure}
    \centering
    \includegraphics[width=0.9\columnwidth]{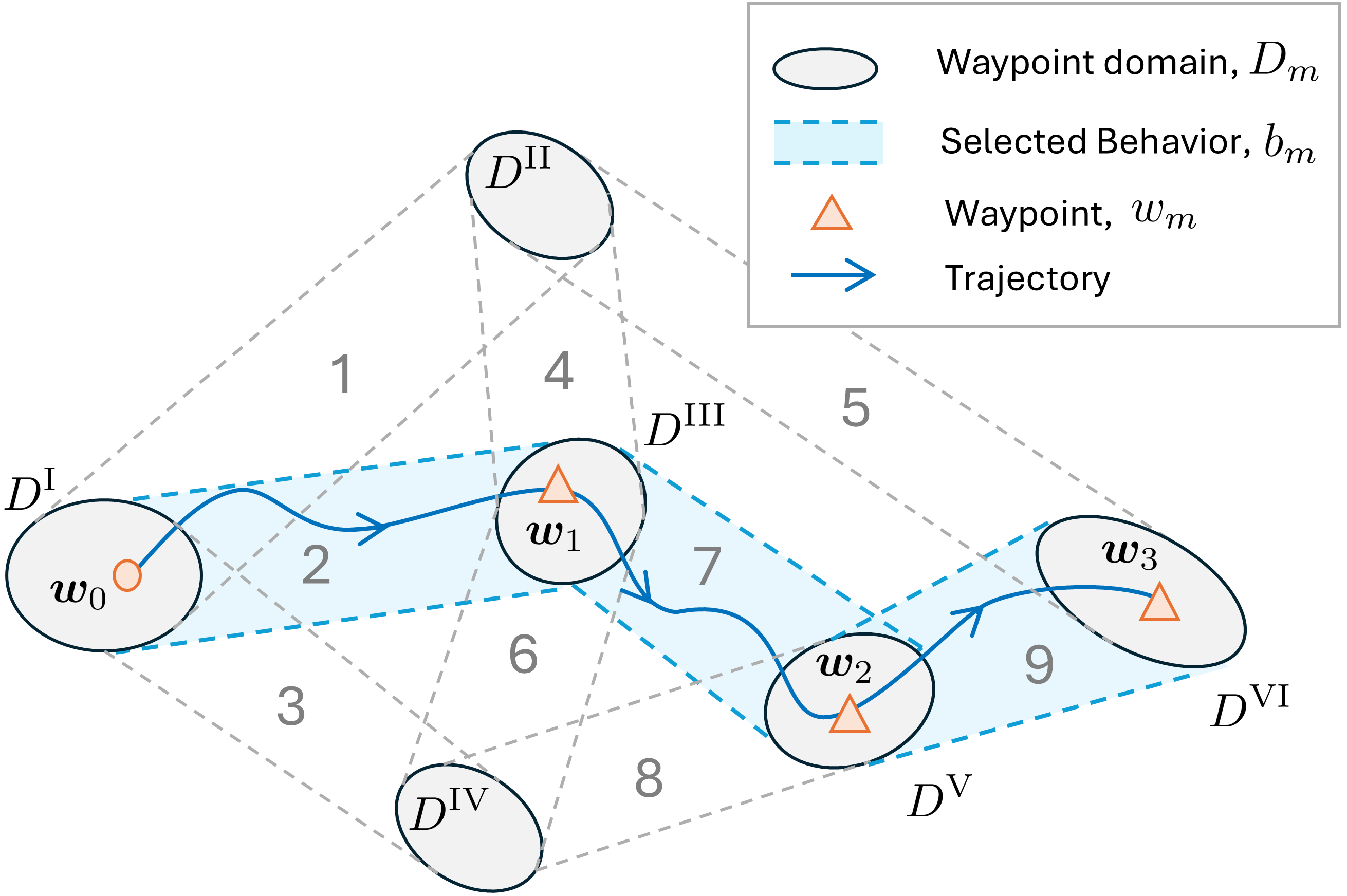}
    \caption{Example behavior--waypoint graph. }
    \label{fig:behavior_graph}
\end{figure}

First, the trajectory is assumed to decompose into a sequence of \textit{behavior primitives}, each representing an admissible transfer between prescribed waypoint domains over an allowable range of transfer times.
Namely, this paper assumes that a trajectory can be represented as a sequence of waypoint-hopping transfers.
This abstraction yields the behavior--waypoint graph $\mathcal{G}(\mathcal{D},\mathcal{E})$, illustrated in Fig.~\ref{fig:behavior_graph}, where $\mathcal{D}$ denotes the set of waypoint-domain nodes and $\mathcal{E} \subseteq \mathcal{D} \times \mathcal{B} \times \mathcal{D}$ denotes the set of behavior-labeled edges, with $\mathcal{B}$ representing the set of behavior labels.
In the highlighted example, the behavior sequence $2\rightarrow 7\rightarrow 9$ induces the domain sequence $D^{\mathrm{I}}\rightarrow D^{\mathrm{III}}\rightarrow D^{\mathrm{V}}\rightarrow D^{\mathrm{VI}}$.
Given the initial state $\boldsymbol{w}_0$, a particular instantiation selects waypoints $\boldsymbol{w}_{1:3}$ from the corresponding domains, yielding a continuous state–control trajectory.
Unlike previous works on kinodynamic motion-planning graphs \cite{starek2017fast, deka2023astrodynamics}, the proposed behavior--waypoint graph operates at the CONOPS-level, representing continuous waypoint domains and admissible behavior transitions rather than discretized states or motion primitives.

\subsubsection{Mission Configuration Interface}

The second key abstraction is \texttt{MissionConfig}, a structured representation of the intermediate mission specification shared between the operator-facing interface and downstream planning modules.
It comprises two fields encoding operator priorities and constraints and four planning fields that are progressively completed throughout the pipeline. These fields are defined below, where $M$ denotes the number of behavior phases and $\bot$ denotes an unspecified entry:
\begin{itemize}
    \item $\mathcal{P}$ (\emph{preferences}): an operator-specified priority ordering over a predefined set of trajectory-quality metrics.
    The metrics are collected in
    $\boldsymbol{q}=[q_1,\ldots,q_{N_{\mathcal{P}}}]^\top$, and $\mathcal{P}$ is represented as a permutation of their indices, with the first index denoting the highest-priority metric. 
    The number of preference metrics is denoted as $N_{\mathcal{P}}$.
    This ordering is used to rank admissible candidates but does not alter their feasibility.

    \item $\mathcal{H}$ (\emph{hard constraints}): operator-specified constraints that must be enforced by the downstream planner.
    Examples include bounds on time of flight, $\Delta v$ budgets, or keep-out zone (KOZ) dimensions.

    \item $\boldsymbol{b}_{1:M}$ (\emph{behavior sequence}): an ordered sequence of behavior primitives selected from the behavior--waypoint graph, where each entry satisfies $b_m \in \mathcal{B}\cup\{\bot\}$ before completion.

    \item $\boldsymbol{d}_{1:M}$ (\emph{duration sequence}): the transfer-time specification associated with the behavior sequence.
    Each entry may be unspecified, fixed, or bounded, i.e.,
    $d_m \in \{\bot\}\cup\mathbb{R}_{+}\cup\{[d_m^{-},d_m^{+}]\}$.
    
    \item $T$ (\emph{total flight time}): either specified by the operator, inferred from the duration sequence, or constrained by an admissible interval, i.e.,
    $T \in \{\bot\}\cup\mathbb{R}_{+}\cup \{[T^{-},T^{+}]\}$.
    
    \item $\boldsymbol{w}_{1:M}$ (\emph{waypoint sequence}): the waypoint specification associated with each behavior primitive.
    A waypoint may be unspecified, specified as a symbolic domain, or specified as a possibly partial relative state,
    $\boldsymbol{w}_m \in \{\bot\}\cup\mathcal{D}\cup(\mathbb{R}\cup\{\bot\})^{6}$.
\end{itemize}
Natural-language commands may express intent at different levels of granularity, ranging from explicit waypoint and timing specifications to qualitative behavioral and trajectory preferences.
The alternative types supported by the duration, total flight time, and waypoint fields allow \texttt{MissionConfig} to preserve explicitly stated intent while marking unresolved decisions with $\bot$ for completion by downstream modules.
This representation also aligns with established RPO operational practice, in which maneuver sequences are organized around predefined waypoints that serve as intermediate operational objectives \cite{kruger2024adaptive}. 

\subsection{Hierarchical Planning}
\label{subsec:hierarchical_inference}

With the two planning representations established, the following subsections describe each stage of the planning pipeline.
The behavior-duration planning and waypoint generation modules expose common completion interfaces without prescribing a particular inference mechanism; therefore, each may be implemented using deterministic or learning-based methods.
Regardless of the implementation, previously populated fields are preserved, and a completed mission specification is accepted only after its dynamical feasibility and safety have been verified through trajectory optimization.

\subsubsection{Intent Parsing}
\label{sec:method:parsing}

The first module, the intent parser, maps the semantics of the input command to an initial \texttt{MissionConfig}.
Using a curated prompt, a pretrained LLM translates the command into a structured JavaScript Object Notation (JSON) representation of the relevant mission semantics, which is subsequently parsed into \texttt{MissionConfig}.
The resulting initial mission specification, $\mathcal{M}_0$, assigns default values to unspecified preferences and hard constraints.
Behavior, duration, and waypoint fields are populated only when directly supported by the command; all unresolved entries remain marked as $\bot$.
An in-depth discussion of the LLM prompting procedure is provided in the Supplementary Material Sec.~C.

\subsubsection{Behavior-Duration Sequence Planning}
\label{sec:method:decision}

The behavior-duration planner determines the sequence of spacecraft behaviors to be executed and the duration of each behavior.
First, the planner enumerates all graph-valid behavior–and-duration plans consistent with the information specified in $\mathcal{M}0$.
Graph validity alone, however, does not guarantee that a candidate admits a feasible or desirable trajectory.
Although each candidate could, in principle, be evaluated by completing its waypoint specification and solving the corresponding trajectory optimization problem, repeating this process for every candidate may be computationally prohibitive.
The evaluator $Q$ therefore approximates this downstream evaluation using domain-specific information:
\begin{align} \label{eq:behav_model}
   [\hat{p}_{\mathrm{conv}}, \hat{\boldsymbol{q}}]
   =
   Q(s,\boldsymbol{b}_{1:M},\boldsymbol{d}_{1:M}),
\end{align}
where $\hat{p}{\mathrm{conv}}$ estimates the probability of trajectory optimization convergence and serves as an empirical proxy for feasibility, while $\hat{\boldsymbol{q}}\in\mathbb{R}^{N\mathcal{P}}$ estimates the mission-performance metrics.
Thus, $Q$ enables candidate evaluation without requiring a separate full-horizon trajectory optimization solve for each plan.
As discussed in Sec.~\ref{sec:case_study}, $Q$ may be instantiated as an optimization-derived trajectory library, a learned surrogate trained on optimization-generated data, or another domain-specific evaluation model.

Formally, each candidate is an instantiated graph walk comprising behavior edges
$(D_{m-1},b_m,d_m,D_m)$, each paired with a transfer duration $d_m$, where $D_m\in\mathcal{D}$ and $b_m\in\mathcal{B}$.
Starting from the known initial domain $D_0$, the planner traverses admissible outgoing edges over the permitted number of phases and assigns each edge a discrete duration selected from its allowable range.
Candidates inconsistent with behavior, waypoint domain, duration, total flight time, or other constraints already specified in $\mathcal M_0$ are discarded.
This procedure yields the finite candidate set $\mathcal{C}(\mathcal{M}0,s)$, whose elements have the form
$c=(\boldsymbol{b}_{1:M},\boldsymbol{d}_{1:M},\boldsymbol{D}_{0:M})$.
Candidates whose predicted convergence probability falls below the threshold $p_{\min}$ are then removed:
\begin{equation}
    \mathcal{C}_{\mathrm{pass}}
    =
    \left\{c\in\mathcal{C}(\mathcal{M}_0,s)
    \mid \hat{p}_{\mathrm{conv}}(c)\geq p_{\min}\right\}.
    \label{eq:behavior_pass_set}
\end{equation}

The remaining candidates are ranked by the deterministic lexicographic map $\rho_{\mathcal{P}}(\hat{\boldsymbol{q}})$ induced by the operator-specified preference ordering $\mathcal{P}$.
Before comparison, the predicted metrics are ordered according to $\mathcal{P}$ and quantized at resolutions
$\boldsymbol{\epsilon}\in\mathbb{R}{\geq 0}^{N{\mathcal{P}}}$.
Consequently, differences within the prescribed resolution for one metric defer the comparison to the next-highest-priority metric.
The selected candidate is therefore:
\begin{align}
    (\boldsymbol{b}_{1:M}^\star,\boldsymbol{d}_{1:M}^\star)
    = \operatorname*{arg\,max}_{(\boldsymbol{b}_{1:M},\boldsymbol{d}_{1:M},\boldsymbol{D}_{0:M}) \in \mathcal{C}_{\mathrm{pass}}}
    \rho_{\mathcal{P}}(\hat{\boldsymbol{q}}),
\end{align}
from which the behavior-duration $(\boldsymbol{b}_{1:M}^\star,\boldsymbol{d}_{1:M}^\star)$ sequence is used to complete $\mathcal{M}_1$ without modifying any entries inherited from $\mathcal{M}_0$.

\subsubsection{Waypoint Generation}

The third layer instantiates the waypoint sequence for the selected behavior-duration sequence $( \boldsymbol{b}_{1:M}^\star, \boldsymbol{d}_{1:M}^\star)$, thereby completing the mission specification and yielding $\mathcal{M}2$.
As with the evaluator $Q$, waypoint generation is problem-specific. The policy $\pi_w$ defines a waypoint-placement strategy that may be implemented through direct optimization, a deterministic geometric or physics-informed heuristic, or a learned predictor:
\begin{equation}
    \widetilde{\boldsymbol{w}}_{1:M}
    =\pi_w(s,\boldsymbol{b}_{1:M}^\star,\boldsymbol{d}_{1:M}^\star).
    \label{eq:waypoint}
\end{equation}
The waypoint module proposes continuous relative-state waypoints $\widetilde{\boldsymbol{w}}_{1:M}$ within the admissible domains $\boldsymbol{D}_{1:M}^{\star}$ associated with the selected behavior primitives.
These proposals are reconciled with any waypoint information supplied by the operator.
If the operator specifies a symbolic waypoint domain, the proposed waypoint is accepted only if it lies within that domain.
If the operator provides a partial relative state, the specified coordinates are preserved, and only the unspecified coordinates are completed.
Waypoint generation therefore completes the partial specification without overwriting operator-provided information; the rule for selecting continuous states within the admissible domains depends on the particular implementation of $\pi_w$.
In the subsequent trajectory optimization, the resulting waypoints may be imposed as fixed constraints or used as warm-start references.

\subsubsection{Trajectory Optimization}

Once \texttt{MissionConfig} is complete, its hard constraints, phase durations, and waypoint specifications are passed to the trajectory optimization module.
This module computes nominal open-loop state and control trajectories subject to the adopted dynamics, waypoint, and safety constraints by solving the following discrete-time optimal control problem:
\begin{subequations} \label{eq:ocp}
\small
\begin{alignat}{2}
    \min_{\{\boldsymbol{x}_k\}_{k=1}^{N},\{\boldsymbol{u}_k\}_{k=1}^{N-1}} \quad &
    \mathcal{J}  \\
    \text{subject to} \quad
    & \boldsymbol{x}_{k+1} = \boldsymbol{f}_k (\boldsymbol{x}_{k}, \boldsymbol{u}_k), \
    && \forall k \in \mathbb{Z}_{1:N-1}, \label{eq:ocp_con_dyn} \\
    & (\boldsymbol{x}_k, \boldsymbol{u}_k) \in \mathcal{S}_k \cap \mathcal{H}_k,
    && \forall k \in \mathbb{Z}_{1:N-1}, \label{eq:ocp_con_path} \\
    & \boldsymbol{x}_1 = \boldsymbol{w}_0, \label{eq:ocp_con_init} \\
    & \boldsymbol{x}_{k_m} \in \mathcal{W}_m, \ && \forall m \in \mathbb{Z}_{1:M}.  \label{eq:ocp_con_wyp}
\end{alignat}
\end{subequations}
Here, $\mathcal{J}$ denotes the objective function, $N$ is the number of trajectory nodes, $\boldsymbol{x}_k$ and $\boldsymbol{u}_k$ denote the state and control input, respectively, $\boldsymbol{f}_k$ represents the discrete-time dynamics, and $\mathbb{Z}_{a:b}= [a,a+1 ..., b]$.
Equation~\eqref{eq:ocp_con_dyn} enforces the system dynamics, while Eq.~\eqref{eq:ocp_con_path} imposes the default safe set $\mathcal{S}_k$ and the augmented constraint set $\mathcal{H}_k$ defined from \texttt{MissionConfig}.
Equation~\eqref{eq:ocp_con_init} defines the initial condition as $\boldsymbol{w}_0$.
In Eq.~\eqref{eq:ocp_con_wyp}, $k_m$ denotes the time step for the $m$-th waypoint, and $\mathcal{W}_m$ is its admissible waypoint domain generated from $\mathcal{M}_2$.

This paper considers two contracts for incorporating the generated waypoints into the optimal control problem.
Under the \textit{variable} contract, $\mathcal{W}_m=D_m^\star$, allowing the waypoint state to vary within its selected admissible domain to improve trajectory feasibility and optimality.
Under the \textit{fixed} contract,
$\mathcal{W}_m={\widetilde{\boldsymbol{w}}_m}$, requiring the trajectory to pass through the waypoint proposed by the waypoint-generation module.

%% file: case_study.tex
\section{Case Study: On-Orbit Inspection in Low Earth Orbit}
\label{sec:case_study}

The proposed architecture is demonstrated in an on-orbit inspection scenario in low Earth orbit, where a servicer spacecraft approaches a target spacecraft and performs close-proximity operations.
Consistent with the hierarchical validation strategy, the case study presents the modules from downstream to upstream: trajectory optimization, waypoint generation and behavior-duration planning, and, finally, intent parsing.
This ordering allows each lower-level module to be established before evaluating its upstream dependencies and, ultimately, the end-to-end pipeline.

The case study considers behavior sequences comprising two or three phases, such that $M\in{2,3}$.
The physical scenario is parameterized by $s=[\boldsymbol{w}_0,\boldsymbol{\alpha}_0,r_{\mathrm{KOZ}}]$, where $\boldsymbol{w}_0$ denotes the initial state of the servicer relative to the target,
$\boldsymbol{\alpha}0=[a,e,i,\Omega,\omega,M_0]$ contains the target spacecraft’s initial Keplerian orbital elements, and $r{\mathrm{KOZ}}$ denotes the radius of the spherical keep-out zone (KOZ) enforced during trajectory optimization.
The relative state is expressed using quasi-nonsingular relative orbital elements (ROE) \cite{damico_phd_2010}, scaled by the target semimajor axis: 
$\boldsymbol{x}=a[\delta a,\delta\lambda,\delta e_x,\delta e_y,\delta i_x,\delta i_y].$

Scenario variation is introduced through the KOZ radius and the target’s initial orbital phase.
Specifically, $r_{\mathrm{KOZ}}$ is selected from ${30,40,50}$~m, and the initial mean anomaly is selected from $M_0\in\{0^\circ,72^\circ,144^\circ,216^\circ,288^\circ\}$. 
The remaining target orbital elements are fixed at $a=6,738.1$~km, $e=5.581\times10^{-4}$, $i=51.6^\circ$, $\Omega=301.0^\circ$, and $\omega=26.2^\circ$. The trajectory is discretized using a time interval of $\Delta t=900$~s with a maximum horizon of $N_{\mathrm{max}}=130$ time steps.

The following subsections define the preference metrics, behavior--waypoint graph, behavior-duration planning and waypoint-generation modules, and intent-parsing procedure used in the case study.

\subsection{Preference Metrics}

In this case study, $\mathcal{P}$ ranks $N_{\mathcal{P}}=4$ metrics: fuel, flight time, effective observation exposure, and safety margin.
Their definitions and the directions of optimality (i.e., minimizing or maximizing) are:
\begin{itemize}
    \item \textbf{Fuel} (minimizing):
    \begin{align}
        q_{\mathrm{fuel}}=\sum_{k=1}^{N-1}\|\boldsymbol{u}_k\|_2,
    \end{align}
    where $\boldsymbol{u}_k$ denotes the impulsive control input.
    \item \textbf{Flight time} (minimizing):
    \begin{align}
        q_{\mathrm{time}}=\sum_{m=1}^{M}d_m.
    \end{align}
    \item \textbf{Effective observation exposure} (maximizing):
    The accumulated proximity-based opportunity proxy is computed as 
    \begin{align}
        q_{\mathrm{obs}}=\frac{\Delta t}{2} \sum_{k=1}^{N-1} \left[g(\rho_k)+g(\rho_{k+1})\right],
    \end{align}
    where $\rho_k$ is the inter-satellite range at node $k$, and 
    \begin{align}
      g(\rho)=\max\left\{0,1-\frac{|\rho-(r_{\mathrm{KOZ}}+\bar{\rho})|}{\bar{\rho}}\right\},
    \end{align}
    where $\bar{\rho}=25$~m in this case study.
    \item \textbf{Safety margin} (maximizing): \begin{align}
        q_{\mathrm{safe}}=\min_k\rho_k-r_{\mathrm{KOZ}}.
    \end{align}
\end{itemize}

\vspace{-0.5em}
\subsection{Behavior-Waypoint Graph}
Table~\ref{tab:waypoint_domains} enumerates the waypoint domain defined in the ROE space.
For all waypoints, $\delta a=0$ is imposed to eliminate secular drift in the along-track direction.
In addition, $\delta e_x=\delta i_x=0$ is imposed to mitigate secular drift from the $J_2$-perturbation while maintaining radial--normal (RN)-plane separation through eccentricity/inclination (E/I)-vector separation \cite{damico_phd_2010}.
The waypoint space is therefore reduced to $a[\delta\lambda,\delta e_y,\delta i_y]\in\mathbb{R}^3$.
For the E/I-separated domains ($E_0$, $E_+$, and $E_-$), wide RN-plane separation is enforced.
Note that the intervals of $a\delta e_y$ and $a\delta i_y$ instantiate four disconnected subdomains admissible for E/I-vector separation \cite{damico_phd_2010}.
In contrast, flat domains ($F_+$ and $F_-$) are located around the along-track axis (i.e., $\pm$V-bar) with almost no RN-plane separation.
The behavior primitives in Table~\ref{tab:behavior_primitives} define transfers between waypoint domains; given $(D_{m-1},b_m,d_m)$, the destination domain $D_{m}$ is unique.
The corresponding admissible duration ranges are also listed in the table.
The adopted graph is not fully connected because of the operational constraints.
Despite its compact symbolic representation, the graph contains five waypoint domains and 13 behavior-primitive types, which instantiate 21 directed edges and induce 466 graph-valid two- or three-phase paths.
For dataset generation, the three variable components of the initial relative state, $[a\delta\lambda,a\delta e_y,a\delta i_y]$, are sampled within the corresponding waypoint domain (cf. Table~\ref{tab:waypoint_domains}) using a discrete grid for training and continuous uniform sampling for held-out evaluation. 
For each behavior, five duration values are uniformly spaced over the admissible range in Table~\ref{tab:behavior_primitives}.
Combining these paths with the discretized duration choices available in each phase gives $89\times5^2+377\times5^3=49{,}350$ discrete behavior-duration plans before accounting for discrete scenario variation and continuous waypoint domains.

\begin{table}[t]
\centering
\caption{Waypoint domains used in the case study.}
\label{tab:waypoint_domains}
\small
\setlength{\tabcolsep}{2pt}
\renewcommand{\arraystretch}{1.05}
\begin{tabular}{c l c c c}
\hline
Domain & Description & $a\delta\lambda$ [m] & $a\delta e_y$ [m] & $a\delta i_y$ [m] \\
\hline
$E_0$ & E/I-separated, Central & $[-5,5]$ & $\pm[30,70]$ & $\pm[30,70]$ \\
$E_+$ & E/I-separated, $+V$-bar & $[100,250]$ & $\pm[30,70]$ & $\pm[30,70]$ \\
$E_-$ & E/I-separated, $-V$-bar & $[-250,-100]$ & $\pm[30,70]$ & $\pm[30,70]$ \\
$F_+$ & Flat, $+V$-bar & $[100,250]$ & $[-5,5]$ & $[-5,5]$ \\
$F_-$ & Flat, $-V$-bar & $[-250,-100]$ & $[-5,5]$ & $[-5,5]$ \\
\hline
\end{tabular}
\end{table}

\begin{table}[t]
\centering
\caption{Behavior primitives used as directed edges in the behavior-waypoint graph.}
\label{tab:behavior_primitives}
\scriptsize
\setlength{\tabcolsep}{2pt}
\renewcommand{\arraystretch}{1.08}
\resizebox{\columnwidth}{!}{%
\begin{tabular}{c ll c}
\hline
ID & Behavior primitive & Directed transitions & \makecell{Duration\\(orbits)} \\
\hline
1 & Station-keeping & Self-loop at all domains & $[1,3]$ \\
2 & Drift, slow short, $+V$ direction & \makecell[l]{$E_0\rightarrow E_+$, $E_-\rightarrow E_0$} & $[2,4]$ \\
3 & Drift, slow short, $-V$ direction & \makecell[l]{$E_0\rightarrow E_-$, $E_+\rightarrow E_0$} & $[2,4]$ \\
4 & Expand R/N separation & \makecell[l]{$F_+\rightarrow E_+$, $F_-\rightarrow E_-$} & $[1,3]$ \\
5 & Shrink R/N separation & \makecell[l]{$E_+\rightarrow F_+$, $E_-\rightarrow F_-$} & $[1,3]$ \\
6 & Approach from $-V$-bar & $F_-\rightarrow E_0$ & $[1.5,4]$ \\
7 & Retreat to $+V$-bar & $E_0\rightarrow F_+$ & $[1.5,4]$ \\
8 & Approach from $+V$-bar & $F_+\rightarrow E_0$ & $[1.5,4]$ \\
9 & Retreat to $-V$-bar & $E_0\rightarrow F_-$ & $[1.5,4]$ \\
10 & Drift, fast long, $+V$ direction & $F_-\rightarrow F_+$ & $[1.5,3.5]$ \\
11 & Drift, fast long, $-V$ direction & $F_+\rightarrow F_-$ & $[1.5,3.5]$ \\
12 & Drift, slow long, $+V$ direction & $E_-\rightarrow E_+$ & $[4,6]$ \\
13 & Drift, slow long, $-V$ direction & $E_+\rightarrow E_-$ & $[4,6]$ \\
\hline
\end{tabular}
}
\end{table}

\vspace{-1em}

\subsection{Trajectory Optimization}

As a specific instantiation of Eq.~\eqref{eq:ocp}, the following trajectory optimization problem is formulated:
\begin{subequations} \label{eq:opt_nonconvex}
\small
\begin{alignat}{2}
    \min_{\{\boldsymbol{x}_k\}_{k=1}^{N},\{\boldsymbol{u}_k\}_{k=1}^{N-1}}  & \  \sum_{k=1}^{N-1} \|\boldsymbol{u}_k\|_2 \\
    \text{subject to} \quad & \boldsymbol{x}_{k+1} = \Phi_k (\boldsymbol{x}_{k} + \Gamma_k \boldsymbol{u}_k),
    && \ \forall k \in \mathbb{Z}_{1:N-1},  \label{eq:con_dyn} \\
    & \boldsymbol{x}_{k0}^\top S_{kl} \boldsymbol{x}_{k0}  \geq 1,
    && \ \substack{\forall k \in \mathbb{Z}_{1:N-1},\\ \forall l \in \mathbb{Z}_{1:N_s}} \label{eq:con_ps} \\
    & \boldsymbol{x}_1 = \boldsymbol{w}_0, \label{eq:con_boundary} \\
    & \boldsymbol{x}_{k_m} \in \mathcal{W}_m, \quad && \ \forall m \in \mathbb{Z}_{1:M}  \label{eq:con_wyp}  \\
    \text{where} \quad & \Phi_{k} = \Phi(t_{k+1}, t_k), \\
    & \Phi_{kl} = \Phi(t_k + \tau_l, t_k), \\
    & \Psi_{kl} = \Psi(t_k+\tau_l), \\
    & S_{kl} =  \Phi_{kl}^\top \Psi_{kl}^\top P \Psi_{kl} \Phi_{kl}, \\
    & \boldsymbol{x}_{k0} = \boldsymbol{x}_k+\Gamma_k\boldsymbol{u}_k.
\end{alignat}
\end{subequations}
Discrete-time dynamics are expressed in Eq.~\eqref{eq:con_dyn}, where $\Phi(t_{k+1}, t_k) \in \mathbb{R}^{6\times6}$ is the state transition matrix of ROE with the secular $J_2$ effect~\cite{koenig2017new}, and $\Gamma_k = \Gamma(t_k) \in \mathbb{R}^{6\times3}$ is the control input matrix~\cite{damico_phd_2010}, and $\boldsymbol{u}_k$ is an impulsive velocity change in the target's radial--tangential--normal (RTN) frame.
The nonconvex passive safety constraint in Eq.~\eqref{eq:con_ps} is imposed at the prescribed controlled-trajectory nodes $t_k$ and post-failure drift evaluation times $\{\tau_l\}_{l=1}^{N_s}\subset[0,\tau^s]$ after an abrupt loss of control.
\(\Psi(t)\in\mathbb{R}^{6\times6}\) is the first-order mapping from ROE to the relative position and velocity resolved in the RTN frame \cite{damico_phd_2010}.
The matrix $P=\operatorname{diag}[\boldsymbol{I}_3/r_{\mathrm{KOZ}}^2,\boldsymbol{0}_3]\in\mathbb{R}^{6\times6}$ defines the geometry of the spherical KOZ.
The passive safety horizon is set to $\tau^s=1$ orbit and sampled at $N_s=50$ equally spaced timesteps.
Additional considerations such as chance constraints \cite{guffanti_jgcd_2023,takubo2024towards}, and continuous-time safety constraints \cite{elango2025continuous,takubo2026language} can be incorporated without changes in the upstream components; these extensions are outside the scope of this case study.
Overall, this class of minimum-fuel problems with safety and waypoint constraints is generally applicable to a wide range of rendezvous guidance and control.

This paper adopts the SCP algorithm SCvx$\star$ \cite{oguri2023successive} to solve the problem during both dataset generation and deployment.
The hyperparameters of the algorithm are summarized in the Supplementary Material Sec.~F.

\subsection{Behavior-Duration Planning and Waypoint Generation} \label{sec:case:behavior_duration}

The case study considers deterministic and learning-based realizations of the two planning modules introduced in Sec.~\ref{sec:method}. 
For behavior-duration planning, candidate quality is obtained either from an optimization-derived leg library or from a learned metric surrogate $Q_\psi$. 
For waypoint generation, continuous states are selected using either a representative-waypoint heuristic or a learned policy $p_\phi$.
The following subsections describe these realizations.

\subsubsection{Coupled Behavior--Waypoint Realization}
\label{sec:case:deterministic}

A simple heuristic selects representative waypoints at the midpoint of each admissible domain in Table~\ref{tab:waypoint_domains}, despite its suboptimality. 
This reduces the problem to joining precomputed legs between representative waypoints, making candidate screening and ranking fast.
A representative waypoint is placed at the midpoint of each connected component of the E/I-separated domains (i.e., $E_0$, $E_+$, and $E_-$), and at the midpoint of each flat domain (i.e., $F_+$ and $F_-$).

An optimization-derived leg library is constructed offline by solving Eq.~\eqref{eq:opt_nonconvex} for every admissible transfer between representative waypoints, duration grid, target semimajor axis, and KOZ radius considered in the case study.
Each library entry records whether the transfer converges and its associated trajectory-quality metrics.
In particular, the library assumes a circular Keplerian orbit ($e=0$ and no $J_2$ perturbations), reducing the target's orbit parameterization to its semimajor axis instead of the full six-dimensional state.
Note that this assumption could induce deployment error. 
However, this approximation keeps the library compact, with only 2,160 entries across the three KOZ radii at the nominal semimajor axis.

At deployment, each behavior--duration candidate in $\mathcal{C}(\mathcal{M}_0,s)$ is expanded over its admissible representative-waypoint chains. Because the measured initial state generally does not coincide with a representative waypoint, the first leg of each chain, from the current state to the first representative waypoint, is solved online using SCP. All subsequent legs are retrieved from the offline leg library.

A chain is retained only if the online first-leg solve and all required library entries are feasible. 
The corresponding leg-level metrics are then composed to rank them: fuel consumption, flight time, and effective observation exposure are summed across the legs, whereas the safety margin is given by the minimum leg-level value. 

This composition provides an auditable mechanism for candidate screening and ranking. It is distinct from the final full-horizon trajectory optimization, which receives the selected waypoint sequence and generates the final verified trajectory.

\subsubsection{Heuristic Waypoint Generation}

The heuristic policy constructs the waypoint sequence incrementally based on the same representative waypoint set used in the coupled behavior--waypoint realization. 
For an E/I-separated destination domain, the heuristic selects the connected component whose midpoint is closest to the preceding reference state, whereas the along-track coordinate is placed at the midpoint of the destination interval; this preserves the sign combination of $(\delta e_y, \delta i_y)$ at the previous waypoint.  
Flat-domain waypoints are placed at their domain centers.

\subsubsection{Learning-Based Waypoint Generation}
\label{sec:case:learned_waypoint}

The learning-based waypoint policy is the conditional model $p_\phi(\boldsymbol{w}_{1:M}\mid s,\boldsymbol{b}_{1:M},\boldsymbol{d}_{1:M})$, where $\phi$ denotes the model parameters.
The model is supervised using the optimization-derived converged trajectories. 
The output parameterization of the model is carefully designed so that it guarantees that decoded waypoints remain within the graph-prescribed waypoint domains, even for domains that support disconnected subdomains, such as E/I-separated domains (i.e., $E_0$, $E_+$, and $E_-$).
Details on network architecture, parameterization, and loss function are provided in the Supplementary Material Sec.~A.

\subsubsection{Learning-Based Behavior-Duration Planning}
\label{sec:case:learned_behavior}

Learning-based behavior-duration planning uses a surrogate model $Q_\psi$, with parameters $\psi$, to predict the convergence of trajectory optimization and metric outcomes of a candidate behavior-duration sequence.
The training dataset is generated by randomizing the scenario and behavior-duration sequence, applying a specified waypoint-generation policy $\pi_w$, and solving the trajectory optimization.
Although flight time is a deterministic function of the behavior-duration sequence in this case study, the surrogate predicts all four preference metrics for the uniform evaluation of the metric-prediction interface.
At deployment, graph-valid candidates are evaluated in a batched forward pass, candidates below the convergence threshold are removed, and the remainder are ranked using the same deterministic preference map $\rho_{\mathcal{P}}$ to determine the top-ranked behavior-duration plan. 
Details on the architecture, parameterization, and loss function are provided in the Supplementary Material Sec.~B.

\subsection{Intent Parsing}
\label{sec:case:intent_parsing}

The intent parser uses a pretrained off-the-shelf LLM to map an operator command into the partial mission specification $\mathcal{M}_0$ as a structured output.
The prompt provides the behavior--waypoint ontology and output schema, and instructs the model to populate only information explicitly supported by the command, leaving unspecified fields at $\perp$. 
Because populated entries are preserved by downstream planning stages, this conservative parsing rule avoids prematurely completing the mission. 
Details on the prompt development are given in the Supplementary Material Sec.~C.

A central challenge in evaluating intent parsing is establishing an unambiguous reference for the operator's intended $\mathcal{M}_0$.
Evaluation commands are therefore generated \textit{label-first}: a mission intent and corresponding reference specification $\mathcal{M}_0^\star$ are constructed, and then the command is rendered systematically using family-specific templates. 
No language model participates in either label construction or command generation, providing an exactly known reference and avoiding circular evaluation. 
The sampled mission families span maneuver composition, quantitative constraints, preferences, and expert overrides.

Based on the label-first command generation, three 100-command splits are developed: \textit{development}, \textit{held-out instance}, and \textit{held-out structural} to develop the structured prompts for the LLM and to evaluate the intent parsing capability in the following section. 
The development split is used for prompt development and contains commands rendered using the nominal family-specific linguistic templates.
The held-out-instance split contains independently sampled mission specifications and behavior compositions expressed using the same linguistic styles as the development split, thereby evaluating generalization to unseen mission instances without introducing an additional language-distribution shift.
The held-out-structural split similarly remains within the supported mission families and behavior--waypoint ontology, but additionally renders the commands using an unseen terse radio-communication register.
This split therefore evaluates sensitivity to changes in linguistic structure while preserving the underlying planning vocabulary and mission semantics.
The details of the commands' linguistic styles are summarized in the Supplementary Material Sec.~D.

At deployment, the parser may abstain when no valid specification is obtained. 
A graph-validity guard rejects behavior--waypoint combinations inconsistent with the graph before downstream planning.

%% file: results.tex
\section{Results} \label{sec:results}

This section evaluates the proposed framework along three axes: architectural scaffolding, model capability, and test-time computation. 
Each axis corresponds to a design decision that arises when deploying a natural-language interface for spacecraft operations and therefore warrants evaluation beyond the architecture itself.

The first axis concerns how domain knowledge is incorporated into the system. 
Because orbital dynamics, operational constraints, and mission-design conventions may not be reliably encoded within a general-purpose LLM, the proposed framework externalizes this knowledge through a structured mission representation, a behavior--waypoint ontology, and dedicated planning modules.
Thus, the distinct contributions of these components and of the resulting hierarchical decomposition must be evaluated separately. 

The second axis concerns the capability of the underlying language model. 
Candidate backends range from compact models suitable for onboard or resource-constrained deployment to frontier models intended primarily for ground-based planning.
Accordingly, the evaluation must consider not only overall performance but also how the nature of failure changes with model capability, including the violation of the structured interface and misinterpretation of the intent. 

The third axis concerns the allocation of computation at inference time. 
Additional test-time computation devoted to reasoning, verification, or search has been shown to substantially improve performance without requiring model retraining \cite{kwok2026llm}. 
Within the proposed hierarchy, this computation may be assigned to semantic refinement during intent parsing or candidate search during behavior planning.
As they address different failure modes, whether their gains are complementary determines how a fixed computational budget should be distributed across the pipeline.

Together, these axes motivate the following research questions, which are evaluated at the level of both the individual planning modules and the complete language-to-trajectory pipeline:\footnote{All experiments were conducted on a single NVIDIA GeForce RTX 4090 GPU with an AMD Ryzen 9 7950X CPU.}
\begin{enumerate}
    \item \textbf{RQ1 (Architecture)}: What are the distinct contributions of architectural scaffolding and hierarchical planning to language-to-trajectory generation?
    \item \textbf{RQ2 (LLM Capability)}: How does LLM capability affect structured intent recognition, failure modes, and end-to-end mission satisfaction?
    \item \textbf{RQ3 (Test-time Compute)}: How does the additional test-time computation across the planning hierarchy affect intent reliability and realized trajectory preference?
\end{enumerate}

\subsection{Module-Level Validation and Representative End-to-End Realization}
\label{sec:module_validation}

Before addressing the three RQs, this subsection validates candidate realizations of the downstream planning layers and illustrates their integration within the complete architecture. Specifically, alternative waypoint-generation and behavior--duration-planning modules are compared, followed by a representative end-to-end realization showing how the initially partial \texttt{MissionConfig} is progressively completed across the planning hierarchy from a natural-language command.

\subsubsection{Waypoint Generation}

First, waypoint generation is evaluated on 1,000 cases sampled from the held-out test dataset.
No language model or behavior-duration selection is involved in this experiment; only the waypoint proposal and the waypoint constraint handling in the subsequent SCP are varied.
This is the held-out 10\% split of the on-grid dataset in Sec.~\ref{sec:case_study}-\ref{sec:case:behavior_duration}\ref{sec:case:learned_waypoint}, not a continuous-domain test.
The behavior--duration sequence and scenario parameters are fixed for each case.
Three waypoint proposals are compared: a random point in each destination domain, the minimum-displacement heuristic from $\pi_w^{\mathrm{greedy}}$, and the prediction of $p_\phi$.
For the random proposal, a station-keeping phase uses the simple behavior-consistent rule of carrying the preceding waypoint forward rather than sampling a new waypoint.
As discussed in Eq.~\eqref{eq:ocp_con_wyp}, each waypoint proposal is evaluated under two waypoint contracts in the trajectory optimization: \emph{fixed} and \emph{variable}.
Table~\ref{tab:wyp_realization} summarizes downstream SCP success, fuel cost, and iteration count. Under the variable-waypoint contract, both structured waypoint proposals achieve approximately 95\% SCP success and substantially lower fuel cost than random initialization, indicating that the waypoint-generation module provides useful continuous realizations of the selected behavior sequence. 
Under the fixed contract, the minimum-displacement heuristic attains the highest success rate, whereas the learned proposal yields lower fuel among converged cases but a lower success rate.
Based on these results, the minimum-displacement heuristic with variable-waypoint SCP is adopted in the subsequent experiments as a simple, effective, and auditable realization of the waypoint generation interface.

\begin{table*}[t]
\centering
\caption{Waypoint-generation performance on 1,000 held-out test cases; fuel and SCP iterations are the mean $\pm$ standard deviation over the cases in which the SCP converges.}
\label{tab:wyp_realization}
\setlength{\tabcolsep}{4pt}
\renewcommand{\arraystretch}{1.10}
\begin{tabular}{lcc c cc c cc}
\hline
& \multicolumn{2}{c}{Random}
&
& \multicolumn{2}{c}{Heuristic}
&
& \multicolumn{2}{c}{Learned $p_\phi$} \\
\cline{2-3} \cline{5-6} \cline{8-9}
& Fixed & Variable & & Fixed & Variable & & Fixed & Variable  \\
\hline
SCP success [\%]
& 84.0
& 95.0
&
& \textbf{95.7}
& 95.1
&
& 77.1
& 95.6 \\
Fuel [$\mathrm{cm/s}$]
& $15.31 \pm 8.83$
& $11.05 \pm 5.37$
&
& $9.37 \pm 5.42$
& $6.73 \pm 4.19$
&
& $7.00 \pm 4.08$
& $\mathbf{6.65 \pm 4.13}$ \\
SCP iterations
& $6.86 \pm 8.01$
& $9.98 \pm 7.43$
&
& $\mathbf{5.85 \pm 7.25}$
& $8.05 \pm 6.13$
&
& $6.72 \pm 7.27$
& $7.21 \pm 7.12$ \\
\hline
\end{tabular}
\end{table*}

\subsubsection{Behavior-Duration Planning}

Behavior-duration planning is evaluated by whether surrogate-guided ranking identifies useful candidates from the graph-valid set without exhaustive downstream trajectory optimization.
The leg-library and learned surrogate $Q_\psi$ realizations use different waypoint contracts and are therefore evaluated independently rather than compared directly.
For each realization, the structured selector is compared against an offline best-of-64 random-search reference: 64 uniformly sampled graph-valid behavior-duration candidates are evaluated using full-horizon SCP under the same waypoint contract, and the best converged trajectory is retained according to the realized preference.
Although computationally impractical for deployment, this procedure provides a strong reference for random selection under a fixed evaluation budget.

The evaluation considers 100 scenarios.
Under the fixed-waypoint contract, leg-library selection is evaluated against random search using the representative waypoints.
By contrast, under the variable-waypoint contract, $Q_\psi$ selection is evaluated against random search using the minimum-displacement heuristic to initialize variable-waypoint SCP.
Note that the evaluated $Q_\psi$ is the surrogate trained using the minimum-displacement waypoint policy and is therefore assessed under the corresponding variable-waypoint trajectory-optimization configuration.
The scenario parameters are sampled from the case-study domain defined in Sec.~\ref{sec:case_study}.
The lexicographic preference is balanced across four priority orders, with fuel, flight time, observation, and safety margin each serving as the highest-priority metric in 25 scenarios while the remaining metrics retain their default order.

\begin{figure}[t!]
    \centering
    \includegraphics[width=0.95\linewidth]{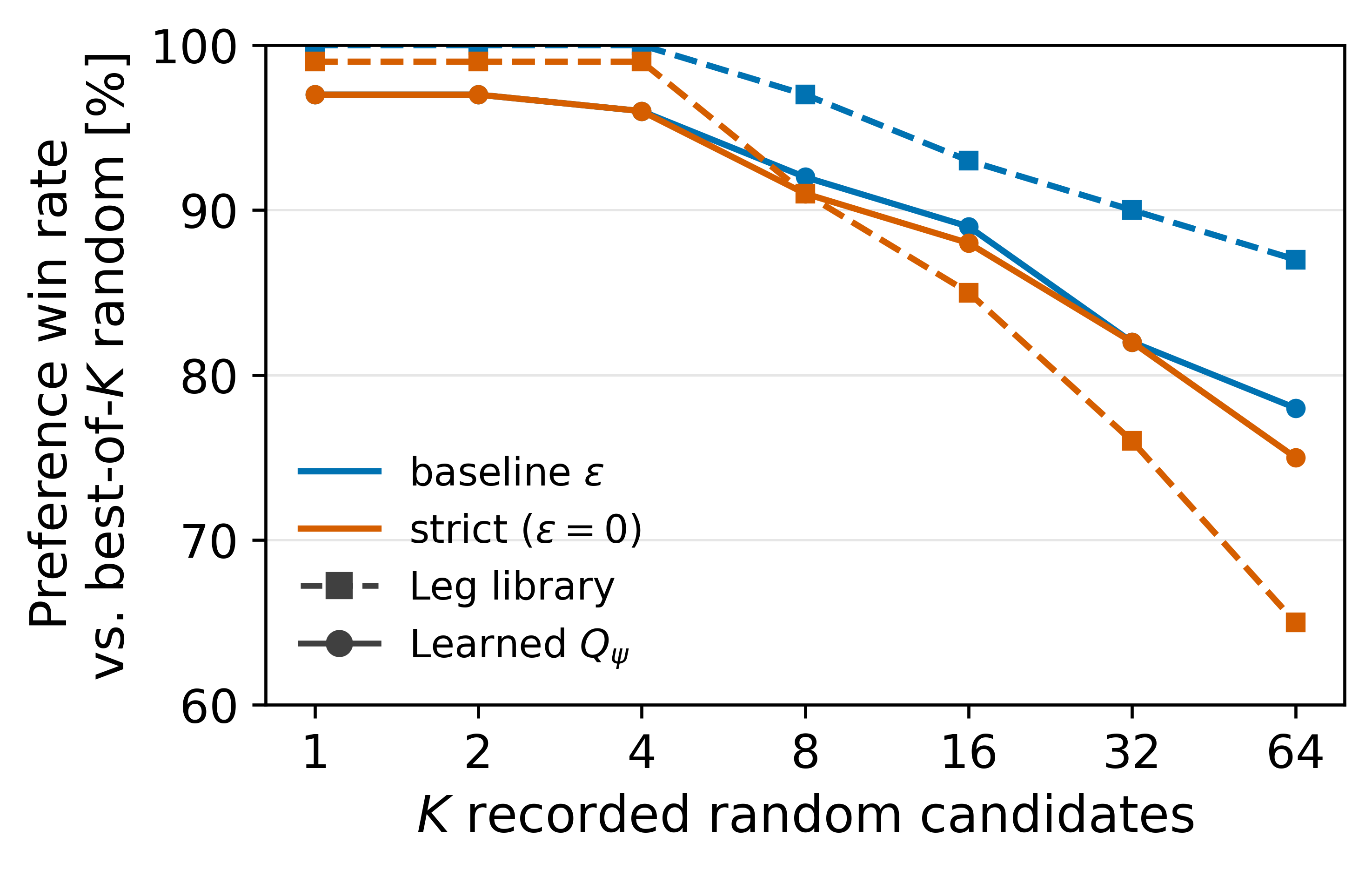}
    \caption{Preference win rate of the structured behavior--duration planning against the best of $K$ uniformly sampled graph-valid candidates over 100 scenarios. }
    \label{fig:q_top1_hit_rate_vs_k}
\end{figure}

Figure~\ref{fig:q_top1_hit_rate_vs_k} reports how often each structured selector outperforms the best-of-$K$ random candidates for $K\leq64$.
For each scenario, the $K$ candidates are sampled from the recorded pool of 64 and ranked under the realized preference.
Sensitivity to the lexicographic margins is evaluated using both strict lexicographic ranking, $\boldsymbol{\epsilon}\simeq\boldsymbol{0}$, and the baseline margin
$\boldsymbol{\epsilon}=[1\,\mathrm{cm/s},900\,\mathrm{s},1\,\mathrm{s},5.0\,\mathrm{m}]$.
Table~\ref{tab:behavior_duration_planning_eval} summarizes the resulting trajectory performance.
Both structured realizations maintain near-universal downstream SCP convergence and select candidates near the top of their corresponding random-search distributions.
The leg-library realization achieves mean candidate percentiles of 98.9\% and 99.6\% under the strict and baseline margins, respectively, while $Q_\psi$ achieves 96.2\% and 96.5\%.
The best-of-64 random baseline is an offline quality reference requiring 64 full-horizon SCP evaluations per scenario, whereas the auxiliary-call column counts only selector-side first-leg SCP solves, of which the leg-library realization requires 60 per scenario (6,000 across 100 scenarios).
Overall, both domain-specific realizations of $Q$ effectively prioritize high-quality behavior-duration candidates without exhaustively solving the full candidate set.

\begin{table*}[t]
\centering
\caption{Behavior-duration planning and corresponding trajectory performance
after trajectory optimization. Fuel and SCP iterations are the mean $\pm$ standard deviation over
successful trajectories.}
\label{tab:behavior_duration_planning_eval}
\setlength{\tabcolsep}{3pt}
\renewcommand{\arraystretch}{1.03}
\begin{tabular}{cccccccccc}
\hline
\multirow{2}{*}{\makecell{Behavior\\selector}}
& \multirow{2}{*}{\makecell{Waypoint\\proposal}}
& \multirow{2}{*}{\makecell{Waypoint\\contract in SCP}}
& \multirow{2}{*}{\makecell{SCP\\success}}
& \multirow{2}{*}{\makecell{Fuel\\$[\mathrm{cm/s}]$}}
& \multirow{2}{*}{\makecell{SCP\\iterations}}
& \multirow{2}{*}{\makecell{Aux. SCP\\calls/request}}
& \multicolumn{2}{c}{\makecell{Percentile \textnormal{[\%]}}}
& \multirow{2}{*}{\makecell{Online\\time \textnormal{[s]}}} \\
\cline{8-9}
& & & & & & & \makecell{Strict} & \makecell{Baseline} & \\
\hline

\makecell{Uniform random\\(best of 64)}
& \multirow{2}{*}{\makecell{Representative\\waypoint}}
& \multirow{2}{*}{Fixed}
& 100/100
& $7.65 \pm 7.04$
& $2.90 \pm 3.80$
& 0
& --
& --
& -- \\

Leg library
&
&
& 100/100
& $5.12 \pm 8.06$
& $3.37 \pm 7.37$
& 60
& 98.9
& 99.6
& 74.77 \\
\hline

\makecell{Uniform random\\(best of 64)}
& \multirow{2}{*}{\makecell{Min.-displacement\\heuristic}}
& \multirow{2}{*}{Variable}
& 100/100
& $3.82 \pm 3.94$
& $4.75 \pm 4.30$
& 0
& --
& --
& -- \\

Learned $Q_\psi$
&
&
& 97/100
& $2.07 \pm 2.79$
& $3.57 \pm 3.93$
& 0
& 96.2
& 96.5
& 3.02 \\
\hline
\end{tabular}
\end{table*}

\subsubsection{Representative End-to-End Realization}

\begin{figure}[t!]
    \centering
    \begin{subfigure}[b]{\columnwidth}
        \centering
        \includegraphics[width=\columnwidth]{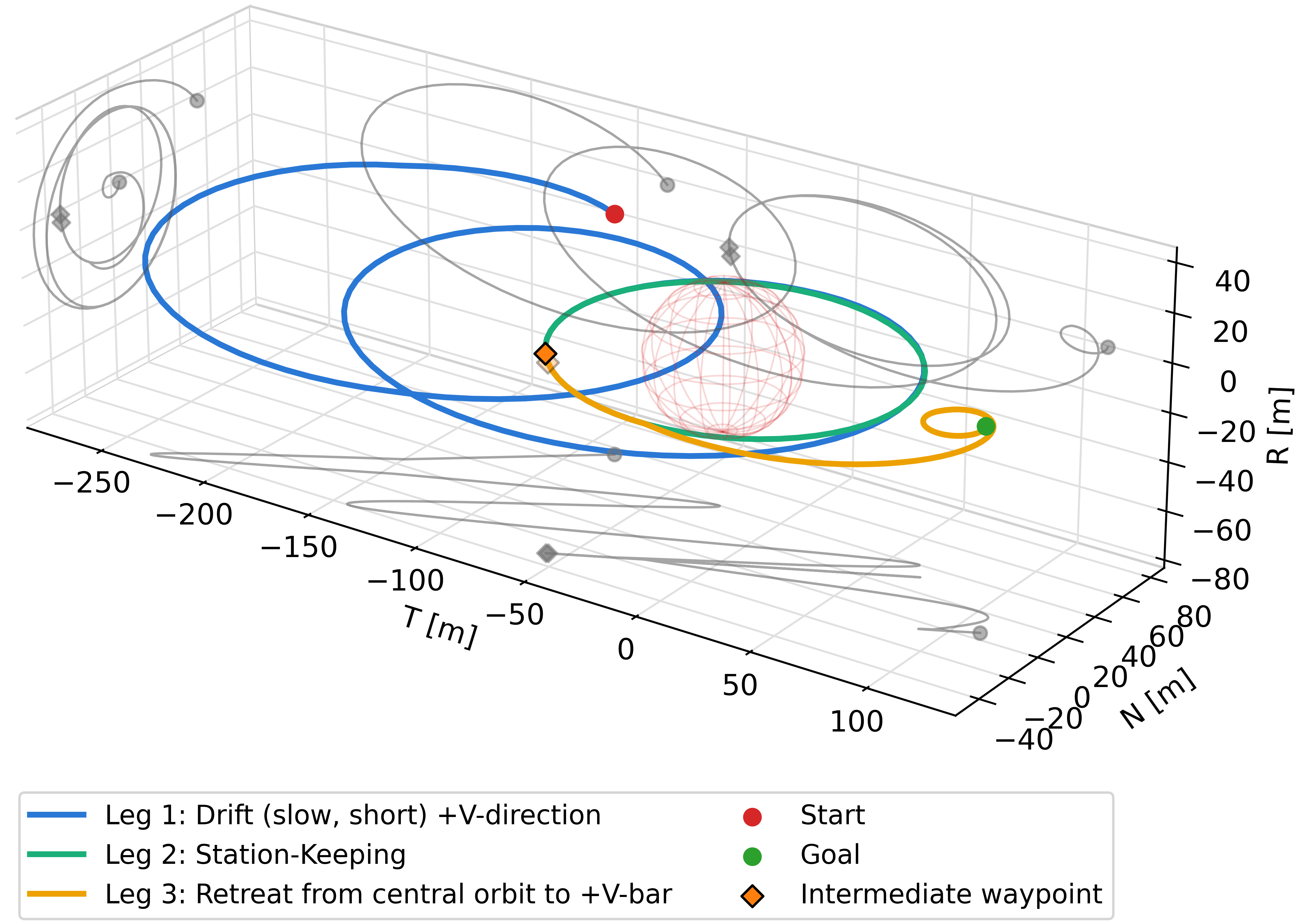}
        \caption{RTN frame}
        \label{fig:sub1}
    \end{subfigure} 
    \\
    \begin{subfigure}[b]{\columnwidth}
        \centering
        \includegraphics[width=0.9\columnwidth]{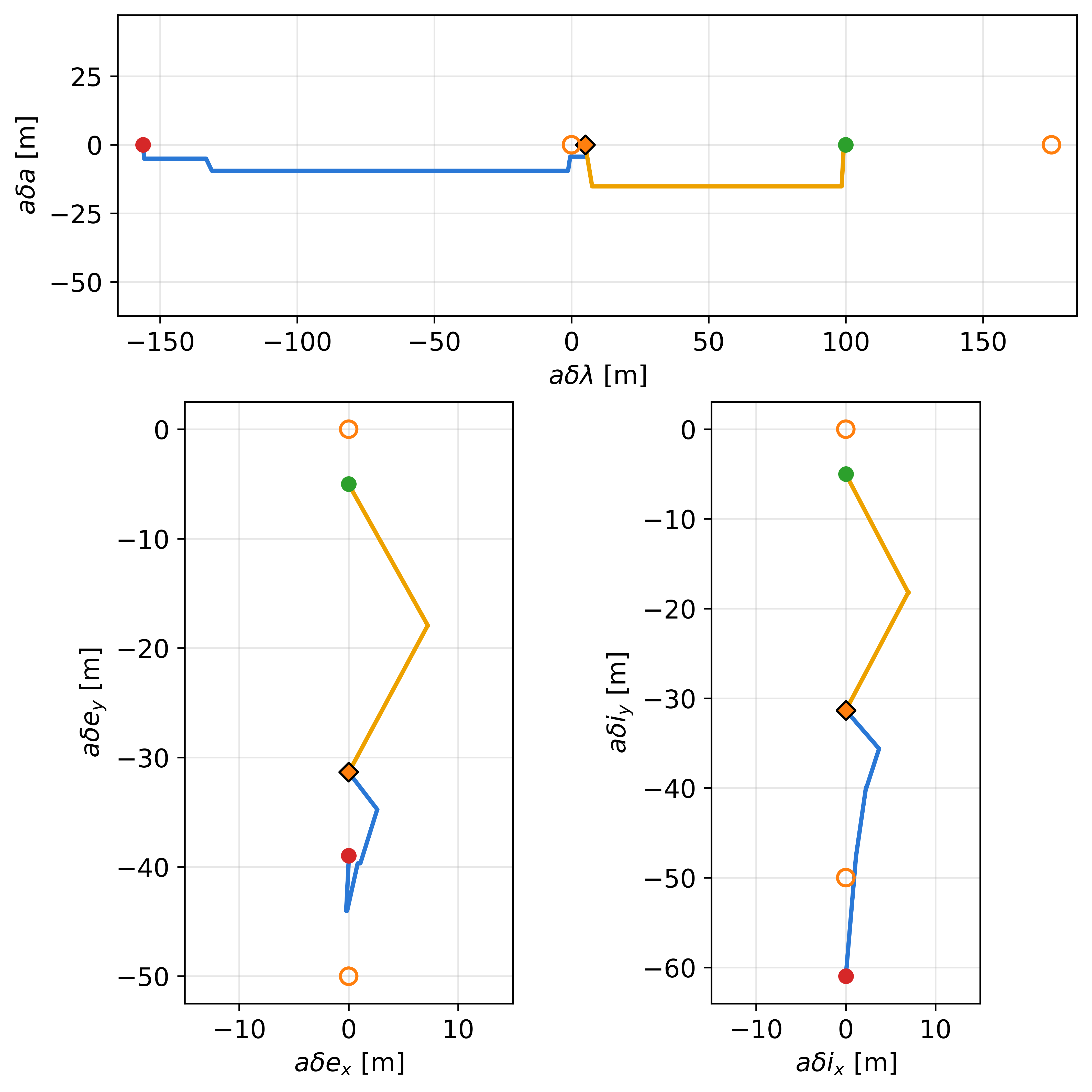}
        \caption{ROE space. Red circles indicate the proposed waypoints used for the warm-starting in the SCP.}
        \label{fig:sub2}
    \end{subfigure}
    \caption{Representative trajectory optimized via SCP based on the completed \texttt{MissionConfig} (cf. Table~\ref{tab:rep_sol_command}).}
    \label{fig:rages_sample_traj}
\end{figure}

Table~\ref{tab:rep_sol_command} traces a representative command through the shared mission representation. 
The command specifies the ordered maneuver structure, the one-orbit near-target phase, the total-time bound, and the priority of fuel over observation, while leaving the remaining phase durations and continuous waypoints unspecified. 
These requirements are preserved in $\mathcal{M}_0$.
The behavior--duration planner completes the sequence with $\boldsymbol{d}=[2.5,1.0,1.5]$ orbits, satisfying the five-orbit bound, and the waypoint module produces the corresponding initialization in $\mathcal{M}_2$.
Full-horizon SCP subsequently adjusts these waypoints while preserving their
behavioral interpretation and enforcing the adopted dynamics and safety
constraints. 
The resulting trajectory is shown in Fig.~\ref{fig:rages_sample_traj}.

\begin{table}[t!]
\centering
\caption{Representative progressive completion of a natural-language command into $\mathcal{M}_0$, $\mathcal{M}_1$, and $\mathcal{M}_2$, followed by the SCP-optimized waypoint sequence.}
\label{tab:rep_sol_command}
\small
\setlength{\tabcolsep}{3pt}
\renewcommand{\arraystretch}{1.02}
\begin{tabular}{c p{0.78\columnwidth}}
\hline
{Entry} & {Representative realization} \\
\hline

Command
&
Drift towards the target, traverse the near-target region for 1 orbit, and
then withdraw toward $+V$-bar. Prioritize minimizing fuel over maximizing
observation quality, and complete the entire operation within 5 orbits.
\\
\hline

Scenario
&
$\boldsymbol{w}_0=[\,0,\;-156.24,\;0,\;-38.97,\;0,\;-60.99\,]$
\\[-1pt]
&
$r_{\mathrm{KOZ}}=30\,\mathrm{m}$ (default), $M_0= 0.0 ^\circ$
\\
\hline

$\mathcal{M}_0$
&
$\mathcal{P}=$[fuel, observation],
\quad
$\mathcal{H}=\emptyset$ (default)
\\[-1pt]
&
$T\in[\bot,\,5.0]~\text{orbits}$
\\[-1pt]
&
$\boldsymbol{b}=[2,1,7]$, \ $\boldsymbol{d}=[\bot,1.0,\bot]~\text{orbits}$, \ $\boldsymbol{w}=[\bot,E_0,F_+]$
\\
\hline

$\mathcal{M}_1$
&
$\boldsymbol{b}=[2,1,7]$
\\[-1pt]
&
(Slow, short drift in the $+V$ direction
\\[-1pt]
&
$\rightarrow$ station keeping $\rightarrow$ retreat to $+V$-bar)
\\[-1pt]
&
$\boldsymbol{d}=[2.5,1.0,1.5]~\text{orbits}$,
\quad
$T=5.0~\text{orbits}$
\\
\hline

$\mathcal{M}_2$
&
Warm-start waypoints [m]:
\\[-1pt]
&
{
\(
\begin{aligned}
\widetilde{\boldsymbol{w}}_{1}
&=[\,0,\;0,\;0,\;-50,\;0,\;-50\,],\\
\widetilde{\boldsymbol{w}}_{2}
&=[\,0,\;0,\;0,\;-50,\;0,\;-50\,],\\
\widetilde{\boldsymbol{w}}_{3}
&=[\,0,\;175,\;0,\;0,\;0,\;0\,].
\end{aligned}
\)
}
\\
\hline

$\boldsymbol{w}^{\star}$
&
SCP-optimized waypoints [m]:
\\[-1pt]
&
{
\(
\begin{aligned}
\boldsymbol{w}^{\star}_{1}
&=[\,0,\;5.00,\;0,\;-31.33,\;0,\;-31.35\,],\\
\boldsymbol{w}^{\star}_{2}
&=[\,0,\;5.00,\;0,\;-31.33,\;0,\;-31.35\,],\\
\boldsymbol{w}^{\star}_{3}
&=[\,0,\;100.00,\;0,\;-5.00,\;0,\;-5.00\,].
\end{aligned}
\)
}
\\
\hline
\end{tabular}
\end{table}

\subsection{RQ1: Effects of Structured Scaffolding}
\label{sec:result_rq1}

This subsection validates the structured scaffolding by performing the architectural ablation of the scaffolded LLM-based TAMP pipeline.
Based on the module-level results in the previous subsection, the subsequent end-to-end experiments adopt the learned surrogate $Q_\psi$ for behavior-duration planning, the minimum-displacement heuristic for waypoint generation, and the variable-waypoint contract for final trajectory optimization, unless otherwise stated.
The complete language-conditioned pipeline is evaluated to determine whether high-level operator intent can be translated into a dynamically feasible trajectory while remaining consistent with the reference mission specification.
Pipeline completion requires the planning process to terminate with a converged SCP trajectory, whereas reference-$\mathcal{M}_0^\star$ satisfaction additionally requires the resulting plan and trajectory to preserve the operator-specified information contained in the reference partial mission specification.

To isolate the effect of planning structure, the proposed hierarchical architecture is compared against two direct-generation baselines using the same GPT-5.6 Terra (high reasoning effort) \cite{openai2026gpt56}.
The proposed architecture follows the full hierarchy described above: the LLM first generates the partial specification $\mathcal{M}_0$, after which $Q_\psi$-based behavior-duration planning and minimum-displacement waypoint generation complete $\mathcal{M}_1$ and $\mathcal{M}_2$.
The \textit{Direct-$\mathcal{M}_2$} baseline is provided with the same \texttt{MissionConfig} schema, behavior--waypoint graph, and physical scenario, but directly generates the completed $\mathcal{M}_2$ in a single LLM stage, thereby bypassing the intermediate behavior-duration search and trajectory-aware candidate evaluation.
The \textit{unscaffolded} baseline receives the operator command and physical scenario, but is not provided with the \texttt{MissionConfig} schema, behavior, waypoint domains, or behavior--waypoint graph; it instead directly proposes numerical durations and ROE waypoints.
For all three architectures, the resulting planning specification is subsequently evaluated through the same full-horizon trajectory optimization module.
The hierarchical architecture is limited to 900 completion tokens per LLM call, whereas the two direct-generation baselines are allowed up to 3,000 completion tokens.

\begin{figure*}
    \centering
    \includegraphics[width=0.8\linewidth]{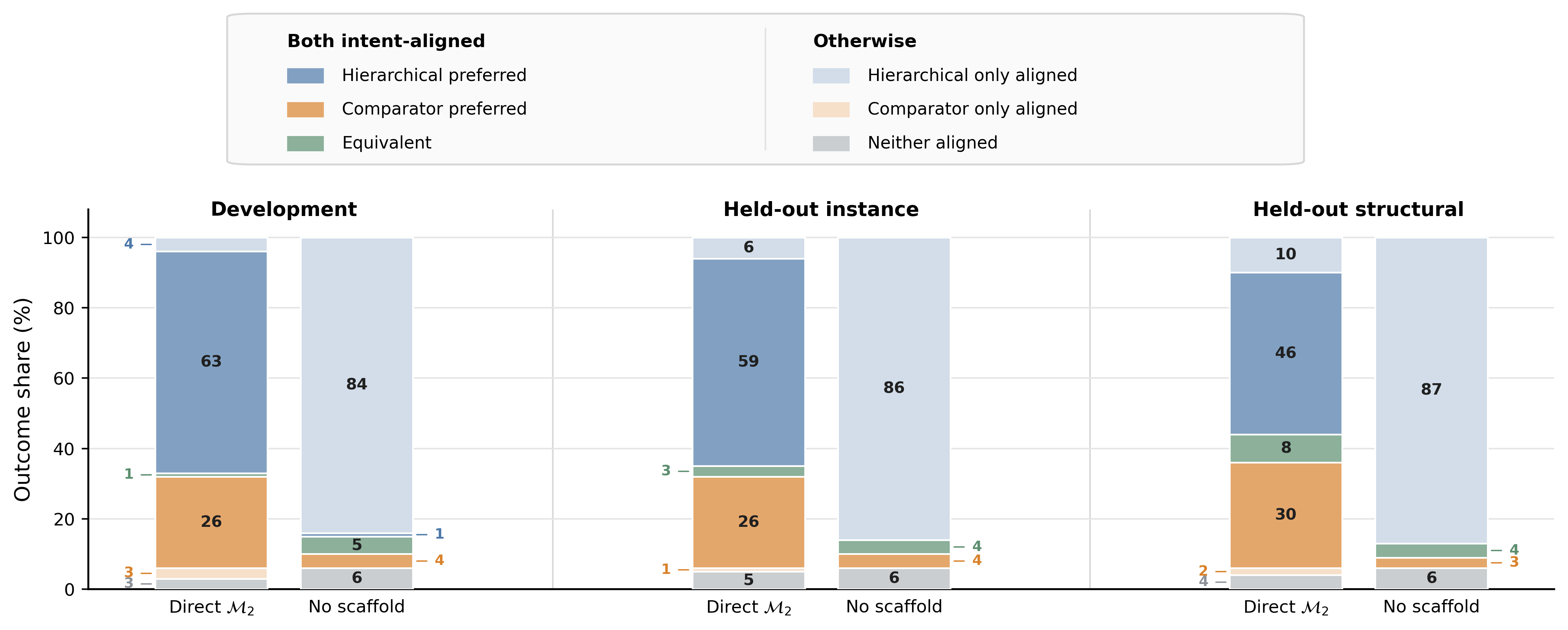}
    \caption{Architecture comparison against the proposed hierarchical architecture, tested with three dataset splits.}
    \label{fig:architecture_comparison}
\end{figure*}

Each baseline is compared pairwise against the proposed hierarchical architecture according to two outcomes: (i) whether each architecture produces a trajectory aligned with the reference mission intent and (ii), when both architectures are intent-aligned, which realized trajectory is preferred under the lexicographic preference mapping.
Accordingly, Fig.~\ref{fig:architecture_comparison} reports the pairwise outcome of each baseline against the proposed hierarchical architecture across the three dataset splits.
The bar chart reports whether both architectures are intent-aligned, only one is intent-aligned, or neither is intent-aligned. 
For cases in which both are intent-aligned, the figure further distinguishes whether the hierarchical trajectory is preferred, the comparator trajectory is preferred, or the two are equivalent under the realized preference mapping.

Across all three datasets, the frontier model rarely produces a realization consistent with the reference mission intent without the operational scaffold.  
Under its fixed 3,000-token generation limit, the unscaffolded planner exhausted the token limit for 45, 44, and 69 commands on the development, held-out-instance, and held-out-structural splits, respectively; these empty responses are counted as operational failures.
In contrast, providing the \texttt{MissionConfig} and behavior--waypoint ontology substantially recovers intent alignment, indicating that the structured representation itself plays an important grounding role rather than serving only as an implementation interface. 
However, when both the proposed architecture and the Direct $\mathcal{M}_2$ baseline produce intent-aligned trajectories, the former realizes the operator's preferred trajectory more frequently in all three dataset splits. 
This result suggests that direct structured generation is often sufficient to identify an admissible mission realization, whereas explicit candidate enumeration and trajectory-aware evaluation provide a mechanism for selecting among multiple admissible alternatives according to feasibility and operator preferences. 
Conversely, the few intent-aligned trajectories produced by the unscaffolded baseline can be competitive in trajectory quality, indicating that its primary deficiency is operational grounding rather than numerical trajectory generation.

\subsection{RQ2: Effect of LLM Capability }
\label{sec:result_rq2}

This subsection examines how the capability of the off-the-shelf LLM backend affects intent parsing and end-to-end trajectory generation.
Three backends spanning two deployment regimes are evaluated: the compact open-weight models Qwen3.5-9B~\cite{qwen3.5} and Ministral3-8B-Instruct~\cite{liu2026ministral}, and the representative frontier model GPT-5.6 Terra with high reasoning effort.
The compact models are loaded using 8-bit quantization, resulting in an approximately \(10\,\mathrm{GB}\) inference memory footprint in the adopted implementation and motivating their consideration for resource-constrained deployment.

Using the label-first benchmark introduced in Sec.~\ref{sec:case_study}-\ref{sec:case:intent_parsing}, the intent-parsing prompt is developed using the 100-command development split and frozen before evaluation.
Performance is reported on this split for reference and on two independent 100-command held-out splits.
The held-out-instance split contains newly sampled mission specifications and behavior compositions expressed using the linguistic styles represented during prompt development.
The held-out-structural split additionally renders the commands using an unseen terse radio-communication register.
Both held-out splits remain within the supported mission families and behavior--waypoint ontology.

The intent-parsing module is evaluated using exact-$\mathcal{M}_0$ match together with atomic precision and recall.
Atomic precision is the fraction of populated atomic fields in the parsed $\mathcal{M}_0$ (including preferences, hard constraints, behaviors, durations, and waypoints) that agree with $\mathcal{M}_0^\star$, whereas atomic recall measures the fraction of reference fields recovered by the parser.

Table~\ref{tab:intent_parsing_eval} shows a clear dependence on LLM capability. 
GPT-5.6 Terra achieves 98\% exact-$\mathcal{M}_0$ recovery on all three splits, whereas Qwen3.5-9B exhibits lower exact recovery and greater sensitivity to the structural split. 
The high atomic precision and recall of the compact models nevertheless indicate that most operator-specified fields are recovered correctly even when the complete specification is not an exact match. 
The larger degradation under the unseen radio-communication register therefore reflects residual semantic sensitivity of the underlying LLM rather than a failure of the structured output interface.

\begin{table}[t]
\centering
\caption{Intent-parsing performance on the prompt-development split and two held-out splits.}
\label{tab:intent_parsing_eval}
\scriptsize
\setlength{\tabcolsep}{5pt}
\renewcommand{\arraystretch}{1.02}
\begin{tabular}{clccc}
\hline
\multirow{2}{*}{\makecell{Dataset\\split}}
& \multirow{2}{*}{Backend}
& \multirow{2}{*}{\makecell{Exact $\mathcal{M}_0$\\{[\%]}}}
& \multicolumn{2}{c}{Atomic match [\%]} \\
\cline{4-5}
& & & Precision & Recall \\
\hline

\multirow{3}{*}{Development}
& Qwen3.5-9B
& 75 & 90.6 & 85.1 \\
& Ministral3-8B-Instruct
& 70 & 88.1 & 89.0 \\
& GPT-5.6 Terra (high)
& 98 & 99.0 & 99.5 \\

\hline

\multirow{3}{*}{\makecell{Held-out\\instance}}
& Qwen3.5-9B
& 78 & 91.1 & 89.8 \\
& Ministral3-8B-Instruct
& 70 & 89.1 & 90.3 \\
& GPT-5.6 Terra (high)
& 98 & 99.5 & 99.0 \\

\hline

\multirow{3}{*}{\makecell{Held-out\\structural}}
& Qwen3.5-9B
& 70 & 91.7 & 82.9 \\
& Ministral3-8B-Instruct
& 71 & 90.1 & 94.5 \\
& GPT-5.6 Terra (high)
& 98 & 98.0 & 98.0 \\

\hline
\end{tabular}
\end{table}

\begin{figure*}
    \centering
    \includegraphics[width=0.8\linewidth]{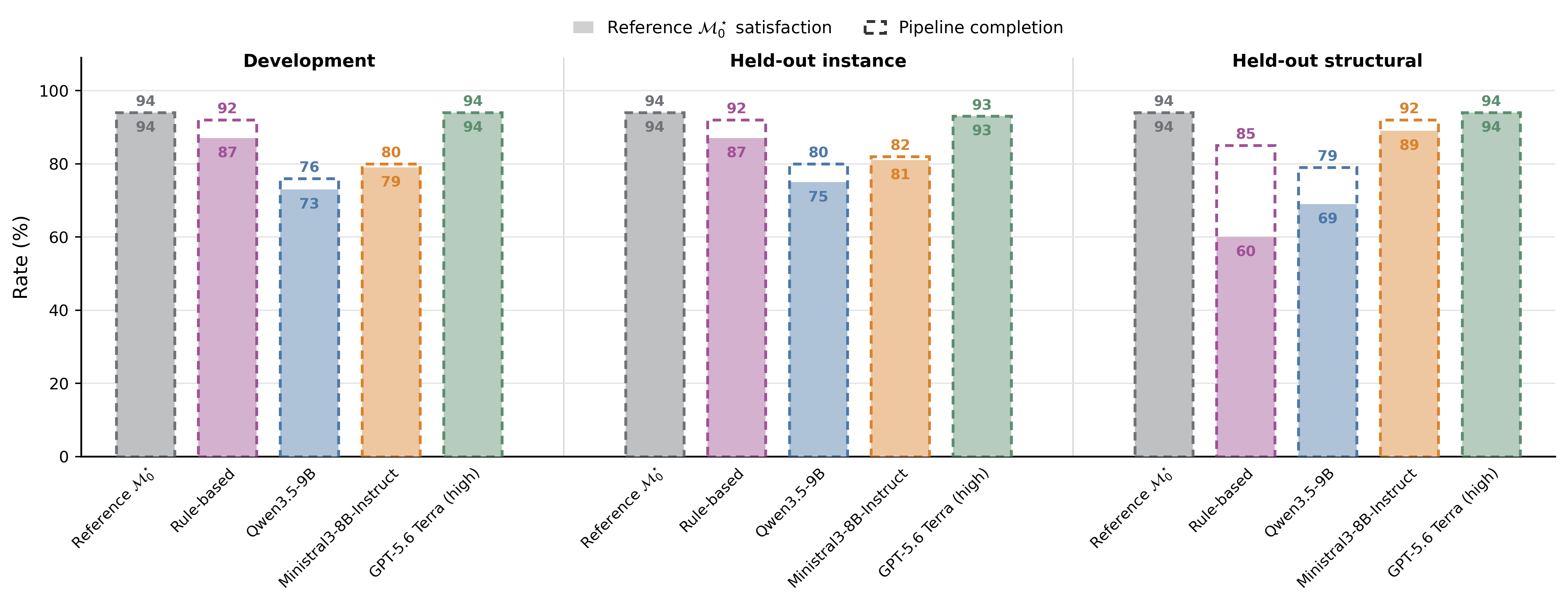}
    \caption{End-to-end mission satisfaction and pipeline completion across semantic parsers on 100 commands per split. All parsed specifications are screened by the same deterministic boundary/path verifier at $N_{\mathrm{rev}}=0$.}
    \label{fig:end_to_end_eval}
\end{figure*}

\begin{figure*}[t!]
    \centering
    \begin{subfigure}[b]{0.95\columnwidth}
        \includegraphics[width=\columnwidth]{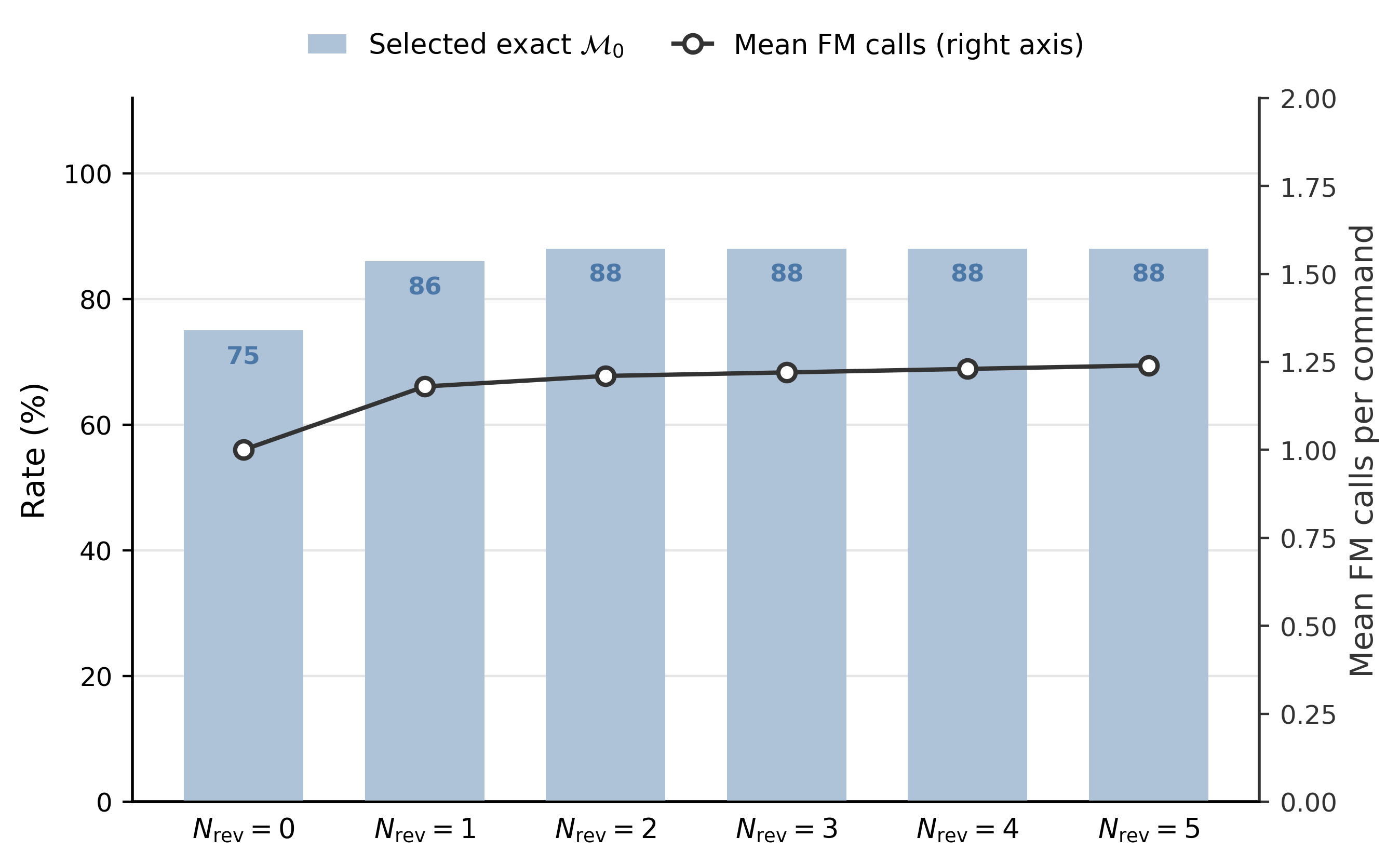}
        \caption{Verifier-guided revision}
        \label{fig:revision_compute}
    \end{subfigure} 
    \begin{subfigure}[b]{1.02\columnwidth}
        \centering
        \includegraphics[width=\columnwidth]{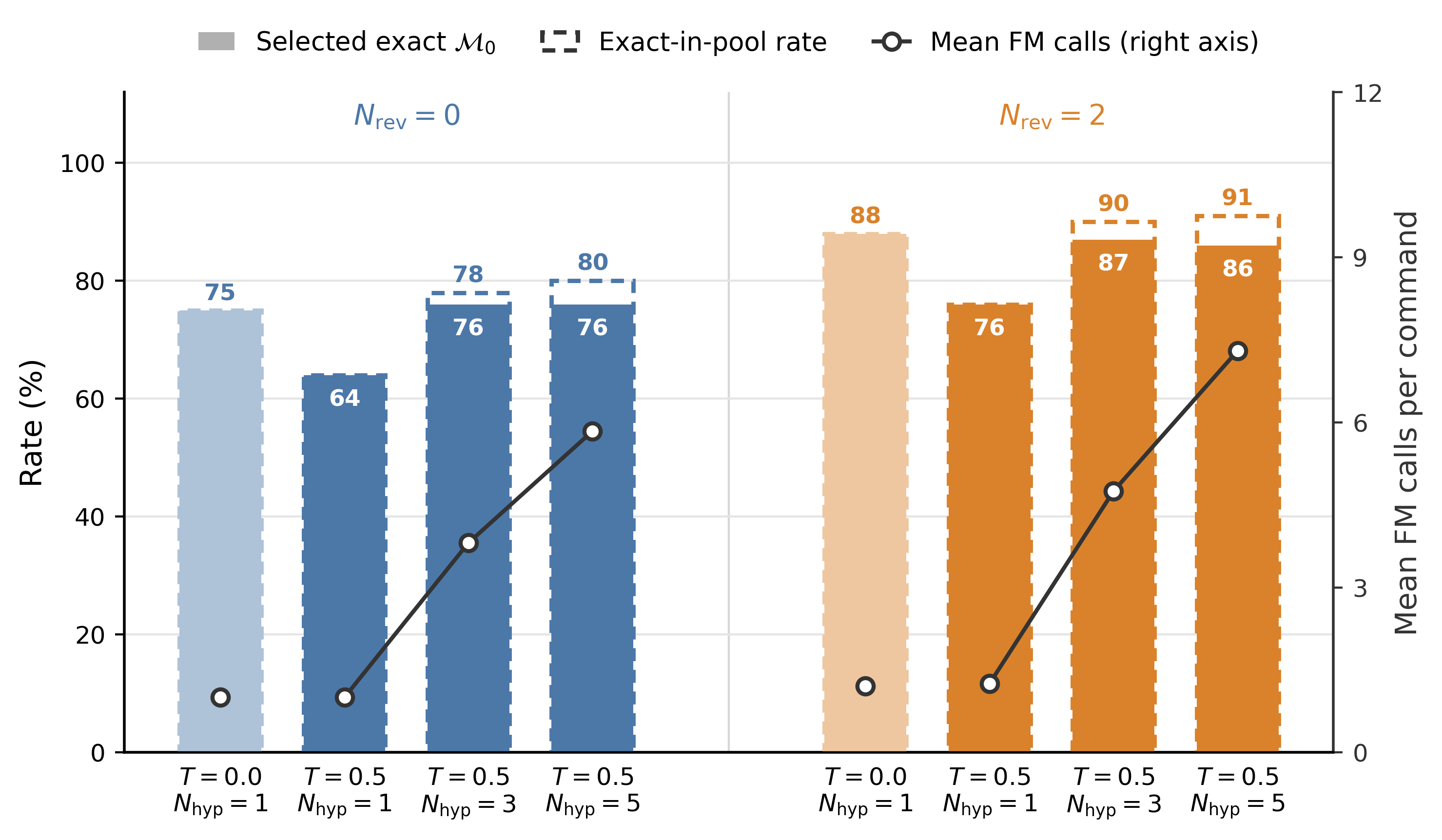}
        \caption{Multi-hypothesis semantic search and selection.}
        \label{fig:intent_compute}
    \end{subfigure}
    \caption{Effect of semantic test-time compute on intent parsing with Qwen3.5-9B and the development-split commands. }
    \label{fig:test_time_compute}
\end{figure*}

\begin{figure}[t!]
    \centering
    \includegraphics[width=\linewidth]{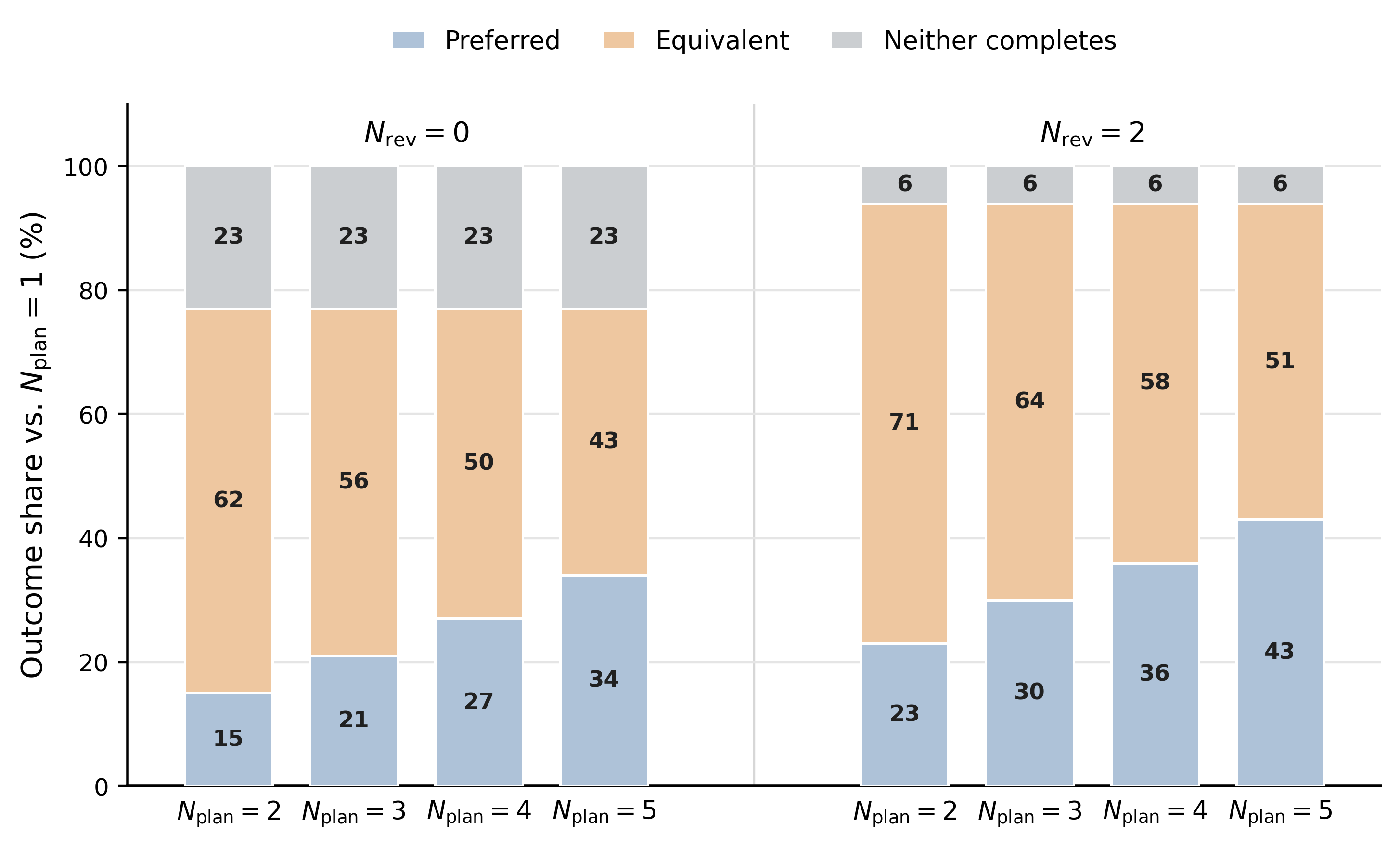}
    \caption{Effect of planning-search budget over the top-$N_\mathrm{plan}$ behavior-duration candidates ranked by $Q_\psi$.}
    \label{fig:plan_search}
\end{figure}

To isolate the contribution of learned semantic parsing, a deterministic, ontology-aware parser is also evaluated.
This baseline uses a fixed maneuver lexicon, ordered clause-matching rules, and regular expressions for preferences and numerical constraints, while leaving unmentioned $\mathcal{M}_0$ fields unspecified.
The rules were developed on the 100-command development split and frozen before held-out evaluation; exact-$\mathcal{M}_0$ recovery is 92\%, 92\%, and 57\% on the development, held-out instance, and held-out structural splits, respectively.
Figure~\ref{fig:end_to_end_eval} evaluates the complete pipeline using the rule-based parser and three language-model parsers.
All downstream planning modules are held fixed to the configuration established in Sec.~\ref{sec:results}-\ref{sec:module_validation}, so only the intent parser is varied.
Every parsed specification is screened by the same deterministic boundary/path verifier used in RQ3; at $N_{\mathrm{rev}}=0$, a detected inconsistency is rejected without an additional model call.
The Qwen3.5-9B development cell reports 73\% reference satisfaction and 76\% pipeline completion using the same $N_{\mathrm{rev}}=0$, $N_{\mathrm{plan}}=1$ artifact as the nominal baseline in Figs.~\ref{fig:test_time_compute} and~\ref{fig:plan_search}.
The figure reports the reference-$\mathcal{M}_0^\star$ satisfaction rate, defined as the fraction of generated trajectories that satisfy the reference mission specification $\mathcal{M}_0^\star$, regardless of whether the intermediate parsed specification exactly matches $\mathcal{M}_0^\star$. 
The pipeline completion rate is also reported as a supplemental measure of downstream execution success.
Passing the reference $\mathcal{M}_0^\star$ directly to the downstream planner provides an empirical ceiling that accounts for failures originating from the behavior-duration, waypoint-generation, and trajectory optimization stages themselves. 
With GPT-5.6 Terra, the language-conditioned pipeline closely approaches this reference across all three command sets, showing that a sufficiently capable semantic backend can recover the operator intent with little additional end-to-end degradation. 
The compact models retain relatively strong performance on commands expressed in the linguistic styles represented during prompt development, but exhibit greater sensitivity to the unseen radio-communication register. 
The rule-based parser achieves reference-$\mathcal{M}_0^\star$ satisfaction and pipeline-completion rates of 87\% and 92\%, respectively, on both the development and instance-level splits, which share semantic frames and differ only in instance values, exceeding both compact models; however, it declines to 60\% and 85\% on the structural split, falling below both compact models in reference satisfaction.
This degradation illustrates that hand-engineered rules can work well within the controlled grammar used for development but remain fragile to changes in linguistic register and phrasing.
The gap between pipeline completion and reference-intent satisfaction is particularly informative in these cases: successful trajectory optimization does not by itself imply that the generated trajectory realizes the commanded mission semantics.

\subsection{RQ3: Test-Time Compute Across the Planning Hierarchy}
\label{sec:result_rq3}

The results in RQ2 show that the frontier LLM approaches the ceiling of the semantic understanding of the proposed framework, whereas smaller models retain a non-negligible error rate in intent parsing. 
The final RQ therefore investigates whether additional allocation of test-time computation can improve performance without retraining or increasing the baseline model size, and how its effect depends on where that computation is allocated within the planning hierarchy.
In this subsection, Qwen3.5-9B is used as a representative compact model, and the 100-command development split is used for the natural-language command data. 
Test-time compute is considered at two levels of the hierarchy: computation at intent parsing through verifier-guided revision and multi-hypothesis generation, and computation at downstream behavior planning through evaluation of additional behavior--duration candidates.

\subsubsection{Intent Parsing}

To provide additional test-time compute at the intent-parsing stage, two methods are considered in this paper.
The first is a verifier-guided revision mechanism, in which each proposed partial mission specification $\mathcal{M}_0$ is checked for compatibility with the required \texttt{MissionConfig} schema, supported ontology, basic structural constraints, and selected movement patterns that can be deterministically inferred from explicitly stated waypoint and maneuver sequences, as detailed in the Supplementary Material Sec.~E.
The verifier is active for every revision budget: at $N_{\mathrm{rev}}=0$, a detected inconsistency causes rejection, whereas for $N_{\mathrm{rev}}>0$, the previous $\mathcal{M}_0$, together with localized verifier feedback, is returned to the same LLM for correction; otherwise, the parser terminates immediately.
Consequently, $N_{\mathrm{rev}}$ specifies a maximum revision budget rather than a fixed number of LLM calls, allowing additional computation to be invoked only when the verifier identifies a correctable inconsistency.

Figure~\ref{fig:revision_compute} shows the resulting trade-off between revision budget and intent-parsing accuracy.
Exact-$\mathcal{M}_0$ recovery increases from 75\% without revision to 88\% at $N_{\mathrm{rev}}=2$, after which no further improvement is observed through $N_{\mathrm{rev}}=5$.
The mean number of LLM calls per command similarly saturates beyond $N_{\mathrm{rev}}=2$, indicating that additional revision opportunities are rarely invoked once the deterministic verifier no longer detects actionable errors.
The remaining errors are therefore primarily semantically incorrect but structurally valid interpretations that are not observable by the verifier.

The second method expands the semantic search by generating multiple candidate mission specifications. 
Specifically, the generation temperature of the LLM, denoted as $T$, is increased to promote diversity; $N_{\mathrm{hyp}}$ candidate $\mathcal{M}_0$ specifications are generated independently, and a final aggregation query selects a single candidate for downstream planning.
In this experiment, greedy decoding $(T=0)$ provides the nominal baseline, while multi-hypothesis generation uses $T=0.5$ to induce diversity of the output. 
The same Qwen3.5-9B backend is used for both hypothesis generation and final aggregation.

Figure~\ref{fig:intent_compute} evaluates this approach for $N_{\mathrm{hyp}}\in\{1,3,5\}$, both without revision and with $N_{\mathrm{rev}}=2$; it also reports the mean LLM calls per command, which includes both hypothesis-generation/revision calls and the final aggregation call.
It is observed that a single stochastic hypothesis is less reliable than greedy decoding, reducing exact-$\mathcal{M}_0$ in both $N_{\mathrm{rev}}=0$ and $N_{\mathrm{rev}}=2$.
Increasing the hypothesis budget to $N_{\mathrm{hyp}}=3$ recovers this loss, reaching 76\% and 87\% exact recovery, respectively. 
However, further increasing the budget to $N_{\mathrm{hyp}}=5$ provides no additional gain in the selected specification despite the increased inference cost.
Importantly, the candidate pool itself continues to improve with additional hypotheses, where the coverage of the exact $\mathcal{M}_0^\star$ (i.e., probability that generated hypotheses contain $\mathcal{M}_0^\star$) increases as the number of the generated $\mathcal{M}_0$ hypotheses increases. 
Thus, stochastic multi-hypothesis generation increases the probability that a correct interpretation is available, but the additional diversity does not necessarily translate into a better final selection.
Beyond a modest hypothesis budget, the limiting factor therefore shifts from candidate generation toward aggregation and selection rather than a lack of candidate coverage.

\subsubsection{Behavior-Duration Planning}

A second axis of test-time computation expands the downstream search over behavior--duration plans for a fixed parsed specification $\mathcal{M}_0$.
Unlike the semantic mechanisms above, this search does not improve intent alignment; instead, it seeks a more preferable trajectory among admissible realizations of the recovered mission specification.
For each $\mathcal{M}_0$, graph-valid behavior--duration candidates are first ranked by the surrogate $Q_\psi$ as described in Sec.~\ref{sec:case_study}-\ref{sec:case:behavior_duration}\ref{sec:case:learned_behavior}.
The top $N_{\mathrm{plan}}$ candidates are then instantiated using the minimum-displacement waypoint heuristic and evaluated through the trajectory optimization under the variable-waypoint contract.
Among the successfully optimized candidates, the final plan is selected according to the realized lexicographic preference.
Thus, increasing $N_{\mathrm{plan}}$ expends additional compute in trajectory optimization rather than additional LLM calls.

Figure~\ref{fig:plan_search} evaluates this planning-search budget relative to the nominal $N_{\mathrm{plan}}=1$ solution.
Because the candidate sets are nested and the best successfully optimized trajectory is retained, increasing $N_{\mathrm{plan}}$ cannot degrade the realized preference relative to $N_{\mathrm{plan}}=1$.
Increasing the search budget progressively improves trajectory quality: evaluating the top five candidates yields a preferred solution for 34\% ($N_{\mathrm{rev}}=0$) and 43\% ($N_{\mathrm{rev}}=2$) of commands relative to using only the top-ranked candidate.
The improvement persists under both revision budgets, indicating that verifier-guided semantic refinement and downstream planning search act on largely distinct aspects of the pipeline.
Together with Fig.~\ref{fig:q_top1_hit_rate_vs_k}, these results show that $Q_\psi$ provides an effective but coarse filter for promising behavior--duration candidates, while full trajectory optimization of multiple top-ranked candidates can correct residual surrogate-ranking errors and improve the final realized preference.
This benefit is therefore complementary to semantic test-time computation, where revision improves the probability of recovering the intended mission specification, whereas behavior-planning search improves trajectory quality after semantic grounding.

%% file: bio.tex
\begin{IEEEbiography}
[{\includegraphics[
    width=1in,
    height=1.25in,
    clip,
    keepaspectratio
]{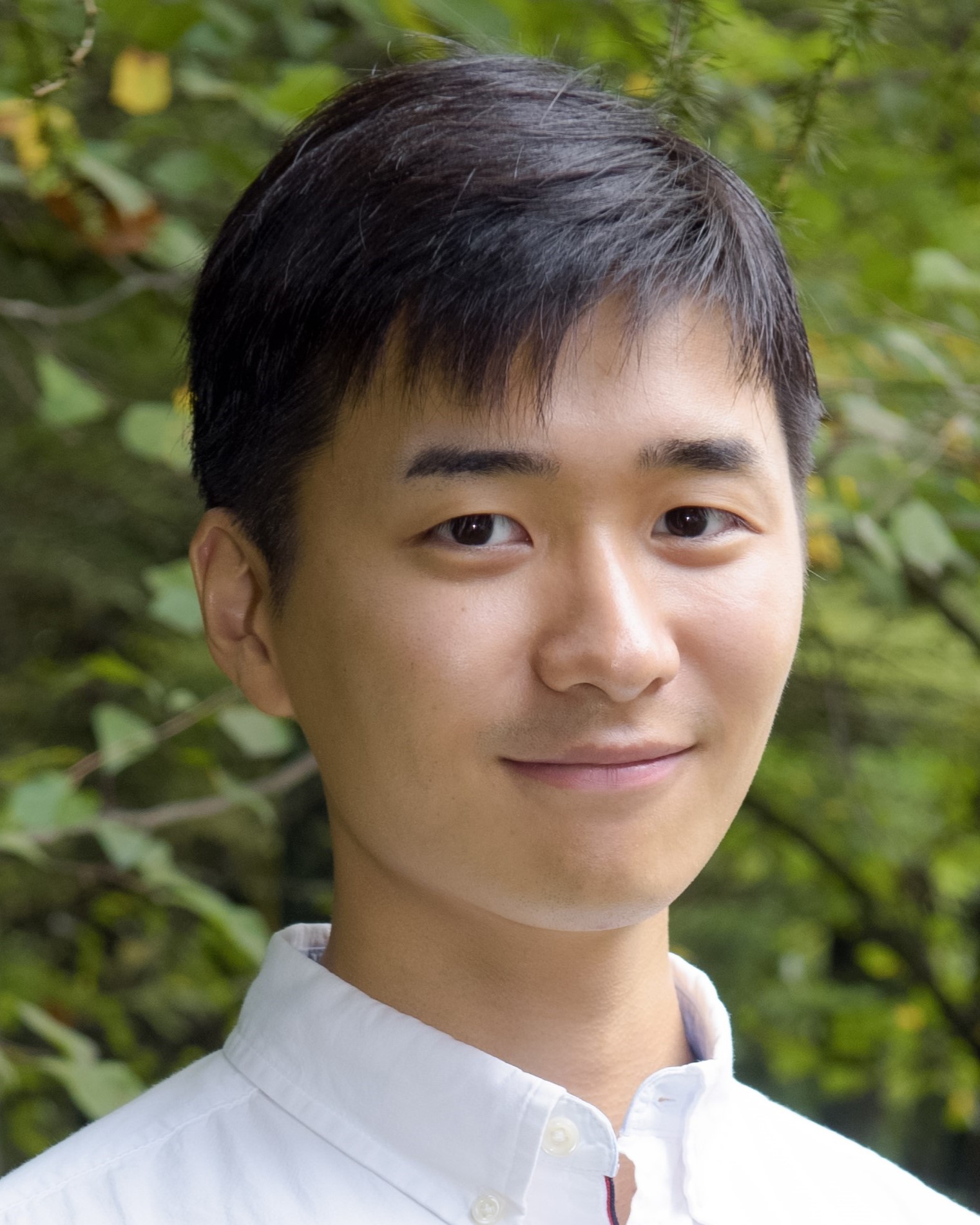}}]
{Yuji Takubo}{\space}(Student Member, IEEE) is a Ph.D. candidate in the Space Rendezvous Lab at Stanford University. He received the M.S. in Aeronautics and Astronautics from Stanford University (2025) and the B.S. in Aerospace Engineering from the Georgia Institute of Technology (2023).
He is a recipient of the NASA Jet Propulsion Laboratory Graduate Fellowship (2023) and the Ezoe Memorial Recruit Foundation Fellowship (2020-).
His research focuses on optimization and astrodynamics, particularly with emphasis on distributed space systems.
\end{IEEEbiography}%

\begin{IEEEbiography}
[{\includegraphics[
    width=1in,
    height=1.25in,
    clip,
    keepaspectratio
]{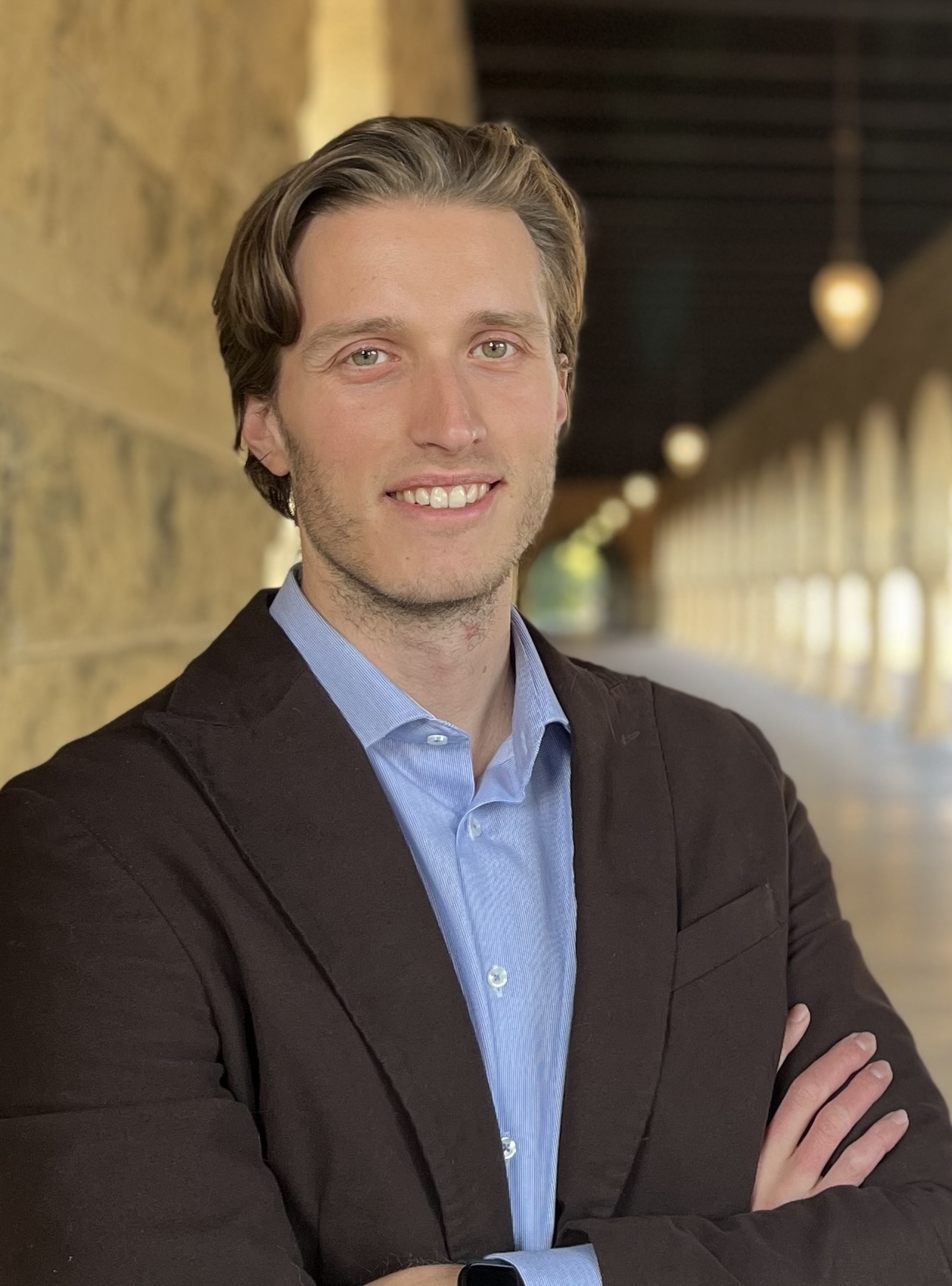}}]
{Daniele Gammelli}{\space} is the Research Director of the Machine Intelligence for Robot Autonomy Laboratory at the Italian Institute of Artificial Intelligence (AI4I), and a Researcher in the Department of Aeronautics and Astronautics at Stanford University. Since 2022, he has been a Research Fellow at the Center for Aerospace Autonomy Research (CAESAR) at Stanford and, until 2025, at the Center for Automotive Research at Stanford (CARS). He received his Ph.D. in Machine Learning and Mathematical Optimization from the Technical University of Denmark (DTU) in 2022, where his doctoral thesis was nominated for the DTU Best Thesis of the Year Award and the DTU Young Researcher Award. Dr. Gammelli’s research focuses on developing the algorithmic foundations and system-level methodologies that enable AI-powered autonomous systems to operate safely, efficiently, and reliably in high-stakes environments, with an emphasis on aerospace autonomy and next-generation mobility systems.
\end{IEEEbiography}

\begin{IEEEbiography}
[{\includegraphics[
    width=1in,
    height=1.25in,
    clip,
    keepaspectratio
]{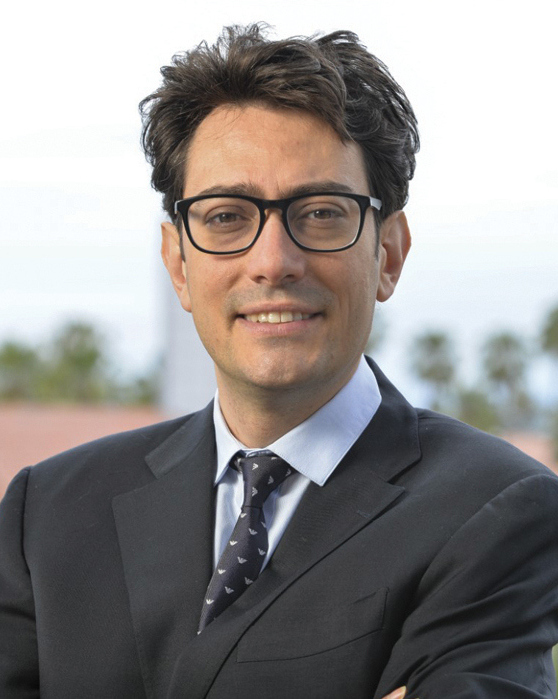}}]
{Marco Pavone} (Member, IEEE) is a Professor of Aeronautics and Astronautics at Stanford University, where he directs the Autonomous Systems Laboratory and the Center for Automotive Research at Stanford. He is also a Distinguished Research Scientist at NVIDIA where he leads autonomous vehicle research. Before joining Stanford, he was a Research Technologist within the Robotics Section at the NASA Jet Propulsion Laboratory. He received a Ph.D. degree in Aeronautics and Astronautics from the Massachusetts Institute of Technology in 2010. His main research interests are in the development of methodologies for the analysis, design, and control of autonomous systems, with an emphasis on self-driving cars, autonomous aerospace vehicles, and future mobility systems. He is a recipient of a number of awards, including a Presidential Early Career Award for Scientists and Engineers from President Barack Obama, an Office of Naval Research Young Investigator Award, a National Science Foundation Early Career (CAREER) Award, a NASA Early Career Faculty Award, and an Early-Career Spotlight Award from the Robotics Science and Systems Foundation. He was identified by the American Society for Engineering Education (ASEE) as one of America's 20 most highly promising investigators under the age of 40. His work has been recognized with best paper nominations or awards at a number of venues, including the European Conference on Computer Vision, the IEEE International Conference on Robotics and Automation, the European Control Conference, the IEEE International Conference on Intelligent Transportation Systems, the Field and Service Robotics Conference, the Robotics: Science and Systems Conference, and the INFORMS Annual Meeting.
\end{IEEEbiography}

\begin{IEEEbiography}
[{\includegraphics[
    width=1in,
    height=1.25in,
    clip,
    keepaspectratio
]{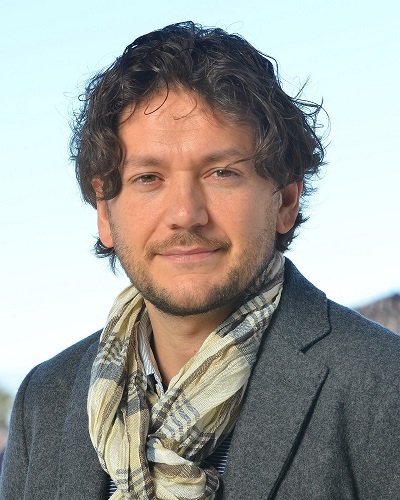}}]
{Simone D'Amico} (Member, IEEE) Associate Professor of Aeronautics and Astronautics (AA) and Professor of Geophysics (by Courtesy). He is the Founding Director of the Stanford Space Rendezvous Laboratory, Director of the Undergraduate Program in AA, and Founding Co-Director of the Center for AEroSpace Autonomy Research (CAESAR) at Stanford. He is a W.M. Keck Faculty Scholar in the School of Engineering and Hoover Science Fellow.
He has 20+ years of experience in research and development of autonomous spacecraft and distributed space systems, including multi-agent architectures such as “rendezvous, proximity operations, and capture”, “satellite formation-flying and swarms”, “fractionated spacecraft”, and “mega-constellations”. He developed the GNC system of several such missions (e.g., TanDEM-X, PRISMA) and is currently the institutional PI of four autonomous satellite swarms funded by NASA and NSF with one of them (NASA Starling) operational in orbit.
Besides academia, Dr. D’Amico is the Chief Science Officer (CSO) of EraDrive. He is in the Advisory Board of four space start-ups focusing on distributed space systems for future applications in SAR remote sensing, orbital lifetime prolongation, and space-based solar power. He received the B.S. and M.S. degrees from Politecnico di Milano (2003) and the Ph.D. degree from Delft University of Technology (2010). Before Stanford, Dr. D’Amico was research scientist and team leader at the German Aerospace Center (DLR). He was the recipient of several awards, including the 2020 IEEE M. Barry Carlton Award, Best Paper Awards at AMOS (2025), IAF (2022), IEEE (2021), AIAA (2021), AAS (2019) conferences, the Leonardo 500 Award by the Leonardo da Vinci Society/ISSNAF (2019), FAI/NAA’s Group Diploma of Honor (2018), DLR’s Sabbatical/Forschungssemester (2012) and Wissenschaft Preis (2006), and NASA’s Group Achievement Awards for the STARLING (2024) and GRACE mission (2004).
\end{IEEEbiography}